\documentclass{article}

\usepackage{PRIMEarxiv}

\usepackage[utf8]{inputenc} 
\usepackage[T1]{fontenc}    
\usepackage{url}            
\usepackage{booktabs}       
\usepackage{amsfonts}       
\usepackage{nicefrac}       
\usepackage{microtype}      
\usepackage{lipsum}
\usepackage{fancyhdr}       
\usepackage{graphicx}       
\usepackage{lineno}
\usepackage{multirow}
\usepackage[numbers]{natbib}
\usepackage{times}
\usepackage{helvet}
\usepackage{courier}
\usepackage{subcaption}
\usepackage{amsmath}
\usepackage{array}
\usepackage{amssymb}
\usepackage{bm}

\usepackage[at]{easylist}
\usepackage{enumitem}
\usepackage{xcolor} 

\newcolumntype{K}{>{\arraybackslash}m{12cm}}
\newcolumntype{L}{>{\arraybackslash}m{9cm}}
\newcolumntype{M}{>{\arraybackslash}m{3cm}}
\newcolumntype{N}{>{\centering\arraybackslash}m{2cm}}
\newcolumntype{Z}{>{\centering\arraybackslash}m{1.5cm}}
\newcolumntype{Y}{>{\centering\arraybackslash}m{1cm}}
\newcolumntype{T}{>{\centering\arraybackslash}m{0.5cm}}

\newcolumntype{O}{>{\arraybackslash}m{6cm}}
\newcolumntype{U}{>{\arraybackslash}m{6.5cm}}
\newcolumntype{P}{>{\arraybackslash}m{4cm}}
\title{Entity-Assisted Language Models for Identifying Check-worthy Sentences}

\author{
  Ting Su, Craig Macdonald, Iadh Ounis \\
  University of Glasgow \\
  Glasgow, UK\\
  \texttt{t.su.2@research.ac.uk, \{craig.macdunald, iadh.ounis\}@glasgow.ac.uk}  \\
}

\begin{document}
\maketitle

\begin{abstract}
We propose a new uniform framework for text classification and ranking that can automate the process of identifying check-worthy sentences in political debates and speech transcripts. Our framework combines the semantic analysis of the sentences, with additional entity embeddings obtained through the identified entities within the sentences. In particular, we analyse the semantic meaning of each sentence using state-of-the-art neural language models such as BERT, ALBERT, and RoBERTa, while embeddings for entities are obtained from knowledge graph (KG) embedding models. Specifically, we instantiate our framework using five different language models, entity embeddings obtained from six different KG embedding models, as well as two combination methods leading to several Entity-Assisted neural language models. We extensively evaluate the effectiveness of our framework using two publicly available datasets from the CLEF’ 2019 \& 2020 CheckThat!\ Labs. Our results show that the neural language models significantly outperform traditional TF.IDF and LSTM methods. In addition, we show that the ALBERT model is consistently the most effective model among all the tested neural language models. Our entity embeddings significantly outperform other existing approaches from the literature that are based on similarity and relatedness scores between the entities in a sentence, when used alongside a KG embedding. 
\end{abstract}


\section{Introduction}
\looseness -1 Politics is an important part of our daily lives. The general public has the right to hold politicians accountable, when they are providing information through debates with their opponents, or when giving speeches. However, general citizens usually do not have enough time to fact check every claim a politician makes, and the vast amount of possible claims made by politicians during their debates and speeches may overwhelm even professional journalists or the fact-checkers of fact-checking organisations (such as Snopes.com and Politifact.org). Indeed, it is common for political debates and speeches to include a mix of factual and non-factual information together, making it possible for the fact-checkers and targeted audiences to be deceived, while making it harder for systems to identify the fake information from the entire debate transcript. To reduce the time and effort required for fact-checking, it is common for fact-checkers and systems to focus their efforts on those dubious claims most likely to matter to the targeted audience. This paper describes an automatic system, that analyses a debate, or a speech, by extracting and ranking all sentences that are worth checking before flagging them to the users by order of their likelihood of being suspicious or fake.

\looseness -1 Recently, the identification of these so-called most {\em check-worthy} sentences has gained increased attention. For example, the ClaimBuster system~\citep{hassan2015detecting} was trained to label sentences in a news article as ``non-factual", ``unimportant factual", or ``check-worthy factual". Moreover, the recent CLEF' 2019 \& 2020 CheckThat!\ Labs~\citep{atanasova2019overview,da2020overview} were introduced as shared evaluation forums where participants were tasked to rank sentences based on their estimated check-worthiness. While CLEF' 2020 CheckThat!\ Lab also includes a task that aims to identify suspicious tweets on Twitter platform, in this work we focus on the debates and speeches. The reasons being tweets have distinct text features compared to speeches and debates, and social network features that speeches and debates do not have. Thus, we consider the identifying check-worthy tweets outside the scope of our task. As is common in recent years, the top-performing participants applied neural language models (LMs), including long short memory models (LSTM) using pre-trained word embedding approaches ~\citep{hansen2019neural,dhar2019hybrid,favano2019theearthisflat}, to represent sentences.

On the other hand, to make a claim is to assert that something is true, while assertion has a common form of  \textit{X verb Y}~\citep{austin1975things}, we therefore define a claim as \textit{conveying $\langle X \ verb \ Y\rangle$ is true}, where the object of the claim is often an entity~\citep{saward2006representative}. 
Moreover, claims made by politicians during debates often contain information about established entities (for instance, entities that are documented in Wikipedia)~\citep{augenstein-etal-2019-multifc}. 
Thus, we focus on established entities in the claim in our check-worthiness identification task, as established entities can be verified with documented information, such as knowledge graphs. 
Knowledge graphs (KGs) are a useful source of information about entities, particularly how they relate to each others. Typically, in a knowledge graph, entities and their relationships  are represented using a triplet structure ($\langle$entity$_1$, relation, entity$_2$ $\rangle$). An example of such a triplet is $\langle$\textit{Arizona}, \textit{a\_state\_of}, \textit{the\_United\_States}$\rangle$. 
Su et al.~\citep{su2019entity} demonstrated that enriching the sentence representation with the similarities and relatedness scores of the entities in that sentence, can significantly improve the identification of the check-worthy sentences than using text representation alone.
Recent works~\citep{yamada2016joint,bordes2013translating,trouillon2016complex} have shown that learned embeddings can be derived from KGs, allowing for the advantages of word embeddings to be applied to the entities found in the KGs. 

\begin{figure}
    \centering
    \includegraphics[width=0.35\textwidth]{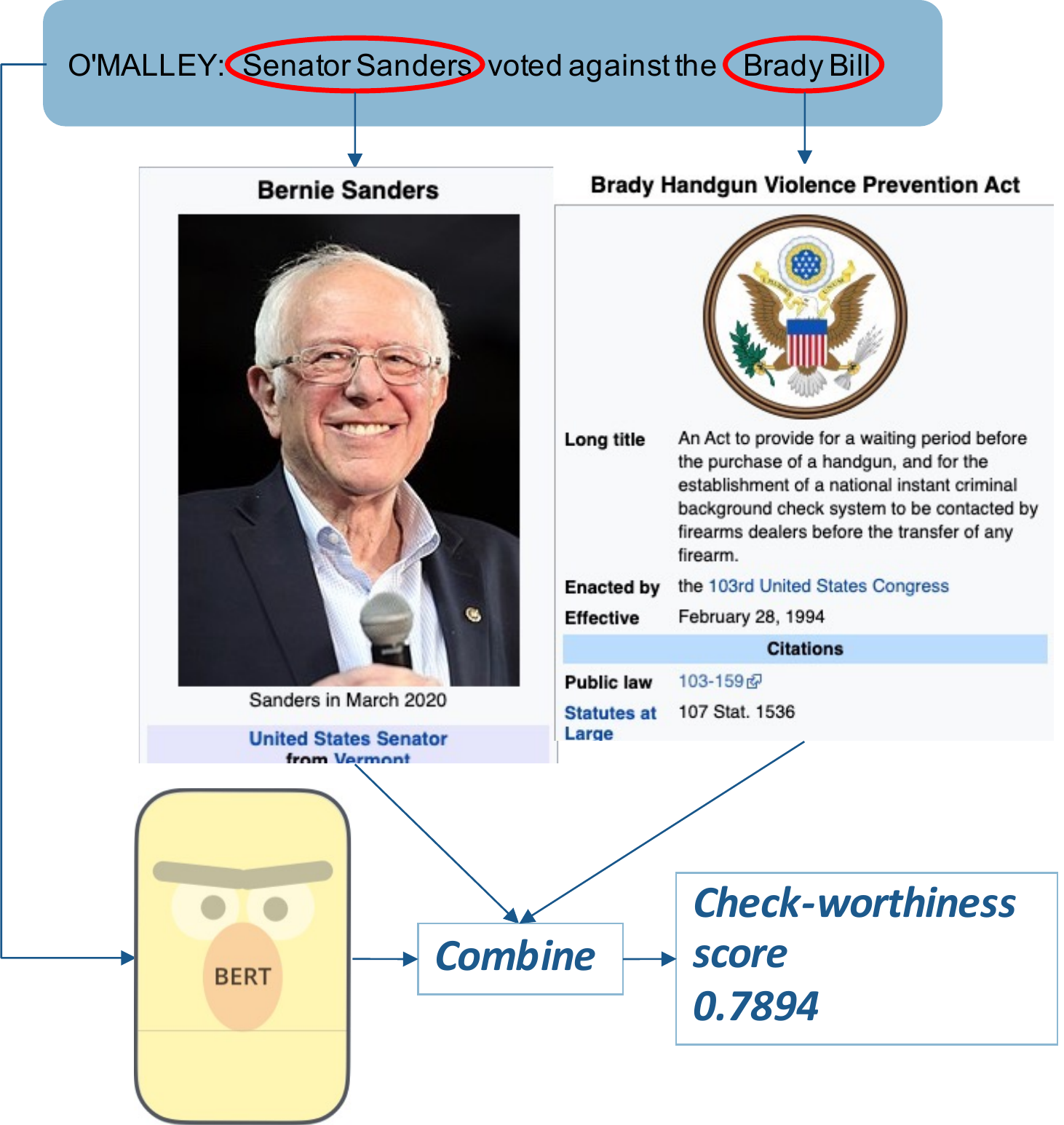}
    \caption{An example showing entities can be informative in identifying check-worthy sentences.}
    \label{fig:motivation}
\end{figure}

Motivated by advances in both language modelling and KG modelling, a recent model, ERNIE~\citep{zhang2019ernie}, demonstrated that by jointly training the language representation and the entity representation, it is possible to leverage the information from both the language representation and entities, thus benefiting a wide range of tasks that require both text and entity information. Building on this work, we hypothesise that the embedded entity vectors obtained from KG embeddings (which we call {\em entity embeddings}) can improve the identification and ranking of check-worthy sentences. For example, Figure~\ref{fig:motivation} demonstrates an example, where the two entities highlight the important components of the sentence, thus helping to identify the sentence as check-worthy. We propose a novel framework, which combines recent neural language models with an entity pair representation for each pair of entities in the sentence, where the entity pair representation is obtained by adequately combining  two  entity  embeddings  extracted  from  an  embedded Knowledge Graph (KG). Our proposed framework allows us to capture rich information about both the language and the entities present in each sentence thereby allowing to better predict their likelihood of being check-worthy. Our framework can be uniformly instantiated to tackle the check-worthiness task either as a text classification or ranking task thereby providing a general and flexible solution for identifying sentences of interest to users. Compared to previous work, our proposed framework has three salient aspects: 
\begin{enumerate}
    \item We represent sentences using language models that go beyond the bag of words and LSTM methods, by leveraging the latest developments in deep neural language models (BERT~\citep{devlin2018bert}, ALBERT~\citep{lan2019albert}, or RoBERTa~\citep{liu2019roberta});
    
    \item We extend the method of incorporating entity information within sentences, from using the simple similarity and relatedness scores between the entities in a sentence~\citep{su2019entity} to a more sophisticated entity representation obtained from KG embeddings. 
    
     \item Our proposed framework does not require the joint training of the language model and the entity representations (for example as done in ERNIE ~\citep{zhang2019ernie}), thereby providing greater flexibility for instantiating and deploying the framework in fact-checking tasks.
    
\end{enumerate}

We instantiate our proposed framework into several Entity-Assisted neural language models, by concatenating the language representation obtained from a state-of-the-art pre-trained language model (e.g. BERT~\citep{devlin2018bert}, ALBERT~\citep{lan2019albert}, or RoBERTa~\citep{liu2019roberta}), with embedded representations for each pair of entities present in a sentence, to represent the sentence's semantic information as well as its entity-related information. Thus, the contributions of this paper are four-fold:
\begin{enumerate}
    \item We propose a simple yet powerful framework to represent sentences with rich entity information, by concatenating together a text model representation with entity pair representations.
    \item Using the CLEF 2019 and 2020 CheckThat!\ Lab datasets, we show that our generated Entity-Assisted neural language models significantly outperform the existing state-of-the-art approaches in the classification task, as well as outperform the participating groups on the CLEF CheckThat!\ leader board in the ranking task. 
    \item We show that representing entity pairs with embeddings is significantly more effective than an existing recent technique from the literature that leverages the similarities and relatedness of the entities. 
    \item Finally, our findings show that among the various knowledge graph embedding models, ComplEx~\citep{trouillon2016complex} leads to the best results, for instance, achieving results as good as the best performing system submitted to the CLEF 2019 CheckThat!\ Lab, without the need for labour-intensive feature engineering.
\end{enumerate}

The rest of the paper is structured as follows: We review the literature related to language representation models, knowledge graph embeddings, as well as the claim check-worthiness task in Section~\ref{sec:lit}. In Section~\ref{sec:model}, we state the task problem, along with our proposed model to address the task. We present our experimental setup in Section~\ref{sec:exp}, and show the results of the experiments in Section~\ref{sec:res}. Finally, we provide concluding remarks in Section~\ref{sec:conclusion}.

\section{Related Work} \label{sec:lit}

In the following, we provide an overview of the key related approaches in language models (Section~\ref{ssLM}), knowledge graph embeddings (Section~\ref{ssKGE})  and check-worthiness identification (Section~\ref{ssCWI}). 

\subsection{Language Models} \label{ssLM}
There are several commonly used methods for representing text, including bag-of-words (BoW) (e.g.\ TF.IDF) \citep{jones1972statistical}, parts-of-speech (POS)~\citep{brill1992simple}, and word embeddings (e.g.\ Word2Vec)~\citep{mikolov2013efficient} representations. 
Building upon word embeddings, deeper long short term memory (LSTM) neural networks (NN), are now commonly used to represent text in a sequential manner, capturing the semantic meaning of the text based on previous tokens. LSTMs encapsulate information about previous tokens, whereas Word2Vec representations are usually based on small skip-grams or windows of tokens.
Such models do not take the future context into account when learning to predict language. 
To address this disadvantage, researchers have developed bidirectional LSTMs (BiLSTMs) and attention-based neural network models (such as BERT~\citep{devlin2018bert}, ALBERT~\citep{lan2019albert}, and RoBERTa~\citep{liu2019roberta}) that not only capture both previous and future tokens in a sentence, but also use the attention mechanism to identify relevant contexts within or between sentences. Moreover, the aforementioned models combine the advantages of a large complex neural network given their pre-training on a large corpus, to create pre-trained neural language models (e.g.\ BERT is trained on Wikipedia, BookCorpus, and Common Crawl~\citep{devlin2018bert}), where the subjective bias from any small training data is minimised. Being based on a state-of-the-art pre-trained language model architecture that can be modified and that has been shown to consistently outperform other models in many tasks~\citep{qu2019bert,yilmaz2019applying}, we use BERT-related language models (i.e., BERT, ALBERT, and RoBERTa) as the base language models in our experiments. 


\subsection{Knowledge Graph Embedding Models} \label{ssKGE}
A knowledge graph (KG) usually contains entities (nodes) and finite types of relationships (different types of edges) between two entities, and can be viewed as a multi-relational graph. 
Each edge in the KG is represented by a triplet $e = \langle e_h, r, e_t \rangle$, indicating that the head entity $e_h$ and tail entity $e_t$ are connected by relation $r$, e.g., $\langle$\textit{Donald\_Trump}, \textit{NomineeOf}, \textit{United\_States\_presidential\_election\_2016}$\rangle$. Such a representation of the structured data is effective at representing factual and trackable relationships, and thus can facilitate the fact-checking processes (e.g., \citep{lin2018fact}).
However, a KG is relatively hard to embed into a lower dimensional vector space.
There are many existing approaches~\citep{bordes2013translating,bordes2014semantic,nickel2011three,wang2014knowledge} that learn embeddings from KGs, by training an unsupervised model based on the co-occurrence of entity pairs and relations. Generally, there are two types of models that are widely used to train KG embeddings: distance-based KG embeddings with ``facts alone'' models~\citep{bordes2013translating,wang2014knowledge,chami2020low} trained on a semantic triplet graph alone (such as FB15k~\citep{bordes2013translating}),
while semantic-based entity embeddings~\citep{bordes2014semantic,nickel2011three} also use the information contained in the corresponding entity descriptions (e.g.\ Wikipedia pages). We describe these two types of models in turn in the next two subsections.

\subsubsection{Facts Alone KG Embedding}
There are two \textit{facts alone} knowledge bases that are widely used in training KG embeddings, namely FB15k and WN18~\citep{bordes2013translating}. The structure of a pure triplet (i.e., a triplet of the form $e = \langle e_h, r, e_t \rangle$, without any additional descriptions for $e_h, r, e_t$) in such knowledge bases enables KG to represent information in a hierarchical and graphical manner. Such representation can be represented in a lower dimension space using graph embeddings, where the learning of the scoring functions is generally based on distances between entities and relationships. Specifically, the Euclidean distance between entities, is used to project the entities based on their relationship with one another, whether these entities and relationships are translated into the same vector space (e.g., TransE~\citep{bordes2013translating}), or into different spaces (e.g., TransR~\citep{linlearning}); or projected into different vector spaces with tensor factorisation (e.g., RESCAL~\citep{nickel2011three}, DistMult~\citep{yang2014embedding}). The advances in deep neural networks also encouraged researchers to deploy deep neutral networks on graph-structured data, such as data encapsulated in a KG. For example, Li and Madden~\citep{li2019cascade} combined a graph embedding method \textit{node2vec}~\citep{grover2016node2vec} with the cascade embedding method, achieving a better performance at predicting triplets, than using TransE alone. Recently, complex space embedding has also been applied to KG embeddings (e.g., ComplEx ~\citep{trouillon2016complex}, RotateE~\citep{sun2019rotate}, QuatE~\citep{zhang2019quaternion}), where the complex valued embedding allows the binary relationship embeddings to represent both symmetrical and asymmetrical relationships (e.g., \textit{$\langle$ Stanley Kubrick, directed, Dr.Strangelove$\rangle$} (asymmetrical) \textbf{cannot} be represented as \textit{$\langle$ Dr.Strangelove, directed, Stanley Kubrick$\rangle$}, while \textit{$\langle$ Barack Obama, married to, Michelle Obama$\rangle$} (symmetrical) can also be represented as \textit{$\langle$ Michelle Obama, married to, Barack Obama$\rangle$}). Finally, the hyperbolic space, with the ability to represent discrete trees in a continuous analogue, has been used for modelling a KG (e.g., MuRP, RotE and RotH~\citep{chami2020low} ), where the multiple possible hierarchical relations of one entity can be modelled simultaneously, resulting in a fewer dimensions of hyperbolic embeddings thereby achieving better performances than those obtained by the Euclidean distance methods.
Aside from the general knowledge bases, specific knowledge bases have been developed to facilitate the KG embeddings of special knowledge. For example, museum information and other unregimented data can be converted into $e = \langle e_h, r, e_t \rangle$ triplets~\citep{pineda2020practical}, while crime-related information can be extracted from newspapers, to build a speciality knowledge base for criminology~\citep{srinivasa2019crime}. 

\subsubsection{Semantic-based KG Embeddings}
Some knowledge bases (e.g., DBpedia) contain more information than just triplets of entities and relationships (e.g. text descriptions for entities, relationships, and their possible features, such as $\langle$\textit{Stanley Kubrick, directed, Dr.Strangelove, a comedy/war movie}$\rangle$)). Hence, a semantic analysis of the available descriptive texts allows algorithms to better capture each entity and its meaning, where a hyperlink between entities serves as a relationship between the two linked entities. To this end, joint training an entity embedding with semantic embeddings can benefit one another. Researchers have explored traditional machine learning methods on jointly trained embeddings, such as random walk~\citep{guo2014robust}, He et al.~\citep{he2013learning} used deep neural networks to compute representations of entities and contexts of mentions from the KB, while Yamada et al.~\citep{yamada2016joint} used a skip-gram method, and trained it on Wikipedia data to obtain the entity embeddings and the associated word embeddings. 

More recently, some researchers (e.g., ERNIE~\citep{zhang2019ernie}, KnowBERT~\citep{peters2019knowledge}) have explored the use of joint training knowledge graph embeddings along with a BERT language model, and showed promising results in several downstream tasks. Bosselut et al.~\citep{Bosselut2019COMETCT} explored whether using the attention mechanism (similar to that for training the BERT model) to enrich a knowledge base embedding with ``common sense knowledge'' embedded in text content is beneficial for more complete KG embeddings. The resulting model, named Comet, puts more emphasis on general information represented as entities (e.g., $\langle$nap, having sub-event, dosing off $\rangle$). 

\subsubsection{Conclusions}
The question of whether the embedded entities are beneficial to suspicious claim detection and/or fake news has not yet been studied. 
For example, the aforementioned Comet model focuses on representing common knowledge instead of entities, which is not suitable for our task. Hence, we do not experiment with this KG embedding model in our present study. However, deploying a joint training of the KB embeddings with the language model may result in less accurate entity embedding information, as well as increase the needed training effort.
Hence, in this paper, we propose to combine KG embeddings with both neural and BoW language representations, to ascertain whether using additional entity information from text enhances the identification of suspicious claims for fact-checking. In the following, we review existing related work about fact-checking and suspicious claim identification.

\subsection{Check-Worthiness} \label{ssCWI}
Identifying check-worthy sentences is a task that identifies the most suspicious and potentially damaging sentences, from a given news article or a political debate that has been divided into sentences. The identified check-worthy sentences should then be prioritised in fact-checking process. 
\looseness -1 The ClaimBuster system~\citep{hassan2015detecting} was the first work to target the assessment of the check-worthiness of sentences. It was trained on data manually labelled as ``non-factual", ``unimportant factual", or ``check-worthy factual", and deployed SVM classifiers with features such as sentiment, TF.IDF, POS, and named entity linking (NEL). Focusing on debates from the US 2016 Presidential Campaign, Gencheva et al.~\citep{gencheva2017context} found that if a sentence is an interruption by one participant in the middle of a long speech by another participant, that was more likely to be selected as check-worthy by at least one news organisation. There are many follow-up works~\citep{patwari2017tathya,jaradat2018claimrank,vasileva2019takes} that have focused on deploying different learning strategies (e.g.\ neural networks, SVM with various features) in reproducing the check-worthy sentences selection process of a news organisation.

\looseness -1 In the 2019 and 2020 editions of the CLEF’2019 CheckThat!\ Lab, datasets of check-worthy sentences from the 2016 US presidential debate were provided -- these were used for 2019 Task 1~\citep{atanasova2019overview} and 2020 Task 5~\citep{da2020overview}.
The top 5 performing groups in the official leader board of the CLEF' 2019 CheckThat!\ Lab are Copenhagen~\citep{hansen2019neural}, TheEarthIsFlat~\citep{favano2019theearthisflat}, IPIAN~\citep{gkasior2019ipipan}, Terrier~\citep{su2019entity}, and UAICS~\citep{coca2019checkthat}.
The 3 groups in the official leader board of the CLEF'2020 CheckThat!\ Lab task 5 are NLP\&IR@UNED~\citep{martinez2020nlp}, UAICS~\citep{cusmuliuc2020uaics}, and TOBB ETU~\citep{kartal2020tobb}.
Table~\ref{tab:clef} provides an overview of the approaches and techniques used by these top performing groups.

\begin{table*}[tb]
\scriptsize
\caption{Methods used by the top five performing groups in CLEF' 2019 \& 2020 CheckThat!\ lab.}
\label{tab:clef}
\centering
\begin{tabular}{|cc|ZTZT|YZTZY|}
\hline
\multicolumn{2}{|c|}{Model}& \multicolumn{4}{c|}{Learning models}                     & \multicolumn{5}{c|}{Features}                \\\hline
 &  & NN (LSTM, Feedforward) & SVM & Logistic regression                & Naive Bayes & word embeddings & syntactic dependence embeddings & SUSE\footnotemark[1]{} & Bag of (NE, POS, Words, n-grams) & hand-crafted features\footnotemark[2]{}  \\\hline
\multirow{5}{*}{2019} & Copenhagen      & \checkmark&           &           &           & \checkmark& \checkmark&           &           &           \\
                      & TheEarthIsFlat  & \checkmark&           &           &           &           &           & \checkmark&           & \checkmark\\
                      & IPIAN           &           &           &           &           & \checkmark&           &           & \checkmark& \checkmark\\
                      & Terrier         &           & \checkmark&           &           &           &           & \checkmark&           &           \\
                      & UAICS           &           &           &           & \checkmark&           &           & \checkmark&           &           \\ \hline 
\multirow{3}{*}{2020} & NLP\&IR@UNED    & \checkmark&           &           &           & \checkmark&           &           &           &           \\
                      & UAICS           &           &           &           & \checkmark&           &           &           & \checkmark&           \\
                      & TOBB ETU P      &           &           & \checkmark&           &           &           &           &           &\checkmark\\\hline    
\end{tabular}
\end{table*}
\footnotetext[1]{Standard Universal Sentence Encoder}
\footnotetext[2]{Readability, sentence context, subjectivity, Sentiment}

Among the groups and systems mentioned in Table~\ref{tab:clef}, the approach deployed by the Terrier group as reported by Su et al.~\citep{su2019entity} is the closest work to our proposed framework in that they also addressed named entity linking, albeit their approach made use of a KG for only the similarity and relatedness between two entities, which we will also be adopting in our experiments. However, Su et al.~\citep{su2019entity} calculated the similarity and relatedness -- in terms of KG structures -- between pairs of entities in the sentences. In contrast, our work here uses recent advances in dense entity embeddings \citep{yamada2016joint,bordes2013translating,linlearning,nickel2011three,yang2014embedding,trouillon2016complex} to provide richer information for suspicious claim identification. We hypothesise that by integrating entity pair representations, we can improve the performance of pure neural language models such as BERT or ALBERT, when identifying check-worthy sentences. It is of note that the best 2019 performing team (Copenhagen) achieved only a mean average precision (MAP) of 0.1660 using the 2019 dataset, while the top 2020 performing team (NLP\&IR@UNED) achieved a MAP of 0.0867, indicating the difficulty of the task. For a fair comparison with existing models on this challenging task, our present work also uses the  CLEF'2019 \& 2020 CheckThat!\ datasets. In the following section we describe the task of check-worthiness prediction, and our proposed entity-assisted language models to address it.

\section{Check-Worthiness Prediction using Entity-Assisted Language Models}
\label{sec:model}

Given a document $d$, we aim to estimate the check-worthiness of each sentence $s_i \in d$. This can be formulated as a classification task, aiming to predict (denoted $\hat{y_i}$) for each sentence if a human would label that sentence as check-worthy or not (c.f.\ $y_i$). 
The task can also be formulated as a ranking task, such that the predicted most check-worthy sentences are ranked highest -- indeed, this is the task formulation taken by the CLEF' 2019 and 2020 CheckThat!\ Labs~\citep{atanasova2019overview,barron2020overview}. In our present study, we propose a uniform framework, which addresses the estimation of check-worthiness both as classification and ranking tasks, when measuring the effectiveness of our models. 

\looseness -1 Our proposed uniform framework for tackling the identification of the check-worthiness of each sentence consists of two components: text representation through the use of language models, and an entity pair\footnote{We also experimented with sentence representation combined with a single entity, and sentence combined with three entities, and neither performs well in this task. For the ease of reading, we do not present the equations and experiments for such structure.} representation obtained from entity embeddings -- discussed further in Section~\ref{sec:structure}. Each sentence is represented by a language model (denoted by $l_{rep}$), which is discussed further in Section~\ref{sec:LM}.
There are three steps involved in representing a pair of entities appearing in a sentence: 
\begin{enumerate}[leftmargin=*]
\item Resolving all entities that appear in each sentence to the corresponding entity using entity linking~\citep{isem2013daiber}; 
\item Transforming the resolved entities into dense entity embeddings through the application of KG embeddings (denoted by $m_{ent}$) -- we discuss the choice of KG embeddings in Section~\ref{sec:ent}; 
\item Each pair of entity embeddings are combined through a combination method (denoted by \bm{$e_{com}$}) to form a single representation for the entity pair. Note that, for sentences that contain more than two entities, every two entities form an entity pair. 
\end{enumerate}

\begin{figure}[tb]
    \centering
    \includegraphics[clip, trim=0.5cm 0cm 0.5cm 0cm, width=0.8\textwidth]{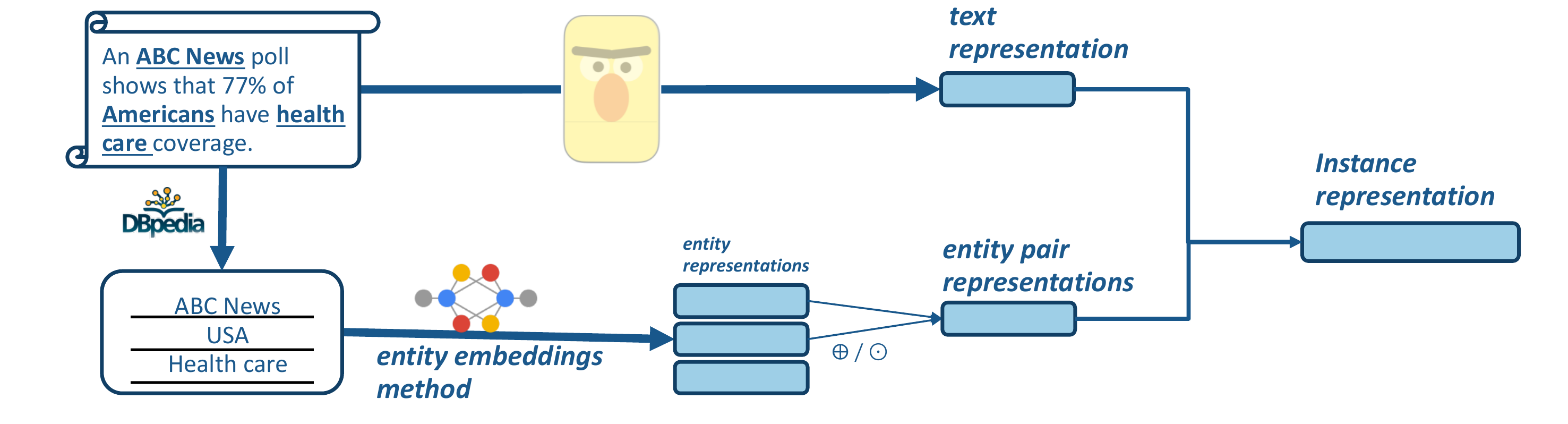}
    \caption{Our proposed Entity-Assisted Language model framework. }
    \label{fig:example}
\end{figure} 

\subsection{Overall Structure of Proposed Framework}
\label{sec:structure}

In order to leverage the semantic representation of various language models, as well as the entities in sentences, for each sentence, we propose to combine its language representation along with an entity pair representation for each pair of entities in the sentence using a neural network framework. 

Firstly, considering a sentence $s_i$ in which a set of entities $E(s_i)$ have been identified through application of an entity linker. Our model is based on pairs of entities, thus we consider {\em input instances} $x_i$, based on pairs of distinct entities:
\begin{equation}
x_i \in \{ \langle s_i, e_h,e_t \rangle \, \forall \, \langle e_h, e_t \rangle \in E(s_i) \times E(s_i) \} \label{eq:instance}
\end{equation}
where $e_h$ and $e_t$ are the head and tail entities. For ease of notation, let $x_i \in s_i$ denote a particular instance $x_i$ obtained from $s_i$ using Equation~\eqref{eq:instance}.
Then, given an input instance $x_i$, we develop two separate models: $f^{cls}(x_i)$ for sentence classification and $f^{rank}(x_i)$ for ranking. Furthermore, for combining the embeddings of a given entity pair, we use two different methods as explained below.

In particular, Figure~\ref{fig:example} shows the architecture of our proposed framework, including the two different methods of entity evidence combination. In the input stage, we use the sentence as input to a language model, so as to obtain the text representation of the input sentence at a semantic level, i.e.:,
\begin{equation}
    l_{rep} = LanguageModel(s_i)
\end{equation}


For \textbf{each entity pair} in an input instance, we represent the relationship (\bm{$e_{rep}$}) between $e_h$ and $e_t$ in a high dimensional space. Thus, we firstly use an existing KG embedding model $m_{ent}$ to extract entity embeddings $\overrightarrow{e_h}$ and $\overrightarrow{e_t}$ for entities $e_h$ and $e_t$. Next, we use a combination method $\bm{e_{com}}$ to obtain the \textbf{entity pair} representation (\bm{$e_{rep}$}). Specifically, as combination methods, we use the element-wise product operation (denoted by \textbf{emb\_prod}), and the concatenation operation (denoted by \textbf{emb\_concat}). This process can be represented as follows: 
\begin{eqnarray}
    \overrightarrow{e_h}=m_{ent}(e_h), 
    \overrightarrow{e_t}=m_{ent}(e_t)\\
    e_{com} \in \{emb\_prod, emb\_concat\}\\
    e_{rep} = e_{com}(\overrightarrow{e_h}, \overrightarrow{e_t})
\end{eqnarray}
\looseness -1 It is of note that we select emb\_prod and emb\_concat because of their wide use as neural operators for combining two vectors (e.g., \citep{cho2014learning,devlin2018bert}, resp.). 


We combine the text representation and entities pair together, to form the input instance representation $x'_i$, by concatenating the language representation ($l_{rep}$) with the entity pair representation ($e_{rep}$):\footnote{We use a uniform $[-1,\ldots,-1]$ vector to represent any entity not having any embedding in the pre-trained KG embeddings.}
\begin{equation}
    x'_i = l_{rep} {\oplus} e_{rep}
\end{equation}

\looseness -1  Next, $x'_i$ can be used both as part of a classification $f_{cls}()$ and ranking $f_{rank}()$ task. In our experiments, $f_{cls}()$ is a fully connected layer with a softmax activation function that estimates the likelihood of each class, while $f_{rank}()$ is a fully connected layer with a sigmoid activation function to obtain the check-worthiness score $\in (0,1)$ for ranking the sentences in descending order:
\begin{eqnarray}
f^{cls}(x'_i) = \frac{e^{((x'_{i}\otimes k) + b)}}{\sum_{j} e^{((x'_{j}\otimes k) + b)}} \label{eqn:cls}\\
f^{rank}(x'_i) =\frac{1}{1+e^{((-x'_i\otimes k) + b)}} \label{eqn:rank}
\end{eqnarray}
where $k$ denotes a fully connected layer kernel and $b$ denotes bias. The objective of our experiments is to identify the most effective $f^{cls}()$ and $f^{rank}()$ models, for classifying and ranking check-worthy sentences, respectively.

Sentences may contain more than one pair of entities, with corresponding different levels of check-worthiness. For these cases, we assume that as long as at least one pair of entities is check-worthy within a sentence, the sentence is check worthy. Thus, the obtained $f^{cls}()$ and $f^{rank}()$ models are applied for each pair of entities in a sentence. Hence, to obtain the final check-worthiness of a given sentence, we take the maximum check-worthiness label/score across all pairs as follows:
\begin{eqnarray}
    \widehat{y^{cls}_i} = \max(f^{cls}(x'_i))\mbox{ }\forall x_i \in s_i \label{eqn:clsall}\\
     \widehat{y^{rank}_i} = \max(f^{rank}(x'_i))\mbox{ } \forall x_i \in s_i \label{eqn:rankall}
\end{eqnarray}
where $x_i \in s_i$ denotes an input instance $x_i$ occurring in sentence $s_i$.

\subsection{Language Models} 
\label{sec:LM}

We aim to understand the robustness of using entity embeddings across a number of language models. In particular, we use a BiLSTM model with an attention mechanism as a representative of non-pretrained language models. We also use several BERT-related neural language models (BERT, ALBERT, RoBERTa) to represent the current state-of-the-art pre-trained language models.  Finally, TF.IDF vectors are used as a representative of traditional BoW models.

\subsection{Obtaining Entity Embeddings from KG Embedding Models}
\label{sec:ent}

\looseness -1 There are multiple ways to analyse entities appearing in the sentence that can benefit the identification of check-worthy material. For example, Ciampaglia et al.~\citep{ciampaglia2015computational} showed that the {\em graph distance} between two entities within a KG (i.e.\ the number of steps on the graph to reach one entity from another) could be used to improve fake news detection accuracy when applying an entity linking method on news articles. On the other hand, Su et al.~\citep{su2019entity} showed that by using additional entity features such as the similarity and relatedness of entity pairs computed using graph distance, an SVM classifier is able to identify check-worthy sentences more accurately than using TF.IDF features alone. However, using graph distance considers only the number of hops between two entities within the knowledge graph, and therefore does not address other possible relationships between the entities (e.g.\ a person (an entity) being \textit{the president} of a country (another entity)). This means that using only the KG's ontology structure results in less information compared to using embeddings that may capture more entity relationships.

Thus, we instead propose to obtain the entity representations using KG embedding models -- such as those introduced in Section~\ref{ssKGE} (e.g.\ \citep{yamada2016joint,bordes2013translating,trouillon2016complex,linlearning,nickel2011three,yang2014embedding}). This allow us to acquire the implicit and hidden KG-based relationships between two entities that are encoded in the embedding vectors that have been learned by a particular model. Indeed, we focus on using pairs of entities, following~\citep{ciampaglia2015computational,su2019entity}.\footnote{Indeed, as we later show in Section~\ref{sec:exp}, sentences containing 2-4 entities are the most frequent in this dataset.}

Different KG embedding models can return varying results when given the same entity and task. For example, Table~\ref{tab:most-similar-examples} shows the four most similar entities for the President of the United States $\langle$\textit{Barack Obama}$\rangle$ obtained using six different KG embedding models that we use in this study. Specifically, Wikipedia2Vec returns the entities that appear closer to the entity \textit{Barack Obama} in the sentence, while the other 5 models show a variety of very specific entities that \textit{Barack Obama} has a relationship with (e.g., the law he passed, the article he wrote, the person he attended the same school with). Such differences in the output provided by the KG embedding models are due to the varying datasets and data structures used to train the models. 

\begin{table*}[tb]
\scriptsize
\caption{Examples of the most similar entities to Barack Obama, using each of the KG embedding models.}
\label{tab:most-similar-examples}
\centering
\begin{tabular}{|c|llll|}
\hline
KG embedding model & \multicolumn{4}{c|}{Most similar entities to \textit{Barack Obama}, in descending order from left to right}\\\hline
Wikipedia2Vec   & Michelle Obama & John McCain & US presidential election & George W. Bush \\
TransE          & Women's History Month & A Child's History of ... & Thickness network ... & Benito Pérez Galdós \\
TransR          & Executive Order 13654	& BODY SIZES OF ...	& ynisca kigomensis & hypothetical protein ...  \\
RESCAL          & Neural representation ... & Natalie Grinczer & Octavia E. Butler & Giuseppe Pozzobonelli  \\
DISTMult        & live preview & Neonatal peripherally ... & KSC - STS-3 Rollout ... & A large sex difference on ... \\
ComplEx        & Peter B. Olney & James Willard Hurst & Robert H. McKercher & William Schwarzer \\\hline
\end{tabular}
\end{table*}

Therefore, the key argument of this paper is that by including the entity embeddings $\overrightarrow{e}$ for each entity $e$ (appearing in the sentence) into our models, we are able to consider the KG-based network relationships of entities in a sentence, when making predictions about the check-worthiness of a sentence. Indeed, entities that are far apart on a simpler word embeddings space may be closer on the entity embedding space, and combining the word embeddings and entity embeddings may be able to bring these two types of information together. Overall, this provides more evidence about the expected co-occurrence of different types of entities within a sentence for identifying those sentences requiring fact-checking.

\section{Experimental Setup}
\label{sec:exp}
Our experiments address the following research questions:
\begin{itemize} [leftmargin=*]
    \item \textbf{RQ1:} Do BERT-related language models outperform the TF.IDF and BiLSTM baselines in identifying check-worthy sentences?
    
    \item \textbf{RQ2:} Does the use of \textbf{entity embeddings} improve the language models' accuracy in identifying check-worthy sentences?
    
    \item \textbf{RQ3:} Which combination method $e_{com}$ performs the best in improving the performance of text representations at identifying check-worthy sentences?
    
    \item \textbf{RQ4:} Which KG embedding model $m_{ent}$ provides entity embeddings that best assists the language models?
\end{itemize}

Moreover, from Section~\ref{sec:model}, the identification of check-worthy sentences can be considered either as a classification task, or instead as a ranking task (as defined by the CLEF' CheckThat!\ Lab organisers). Hence, in the following experiments, we provide conclusions for all RQs from both the classification and ranking perspectives. In the remainder of this section, we describe the experimental setup used to address our four research questions.

\subsection{Dataset}
\looseness -1 All our experiments use both the CLEF'2019 \& 2020 CheckThat!\ datasets. The CLEF'2019 \& 2020 datasets consist of transcripts of  US political debates and speeches in the time period 2016-2019, collected from various news outlets\footnote{ABC, Washington Post, CSPAN, etc~\citep{barron2020overview}, in English only}. Each sentence has been manually compared with \url{factcheck.org} by the organisers. If the sentence appeared in \url{factcheck.org} and is fact checked, it is labelled as a check-worthy claim.
Table~\ref{tab:example} shows a dataset extracted from a speech by Senator Ted Cruz. The CLEF' 2019 \& 2020 CheckThat!\ Labs provided data split for training and testing purposes, which we also use in this paper. Table~\ref{tab:data} shows the statistics of the training and testing sets. In particular, we observe that the prevalence of check-worthy sentences is reduced in the 2020 dataset compared to the 2019 dataset.

\begin{figure}[tb]
    \centering
    \begin{subfigure}[b]{0.63\textwidth}
        \includegraphics[width=\textwidth]{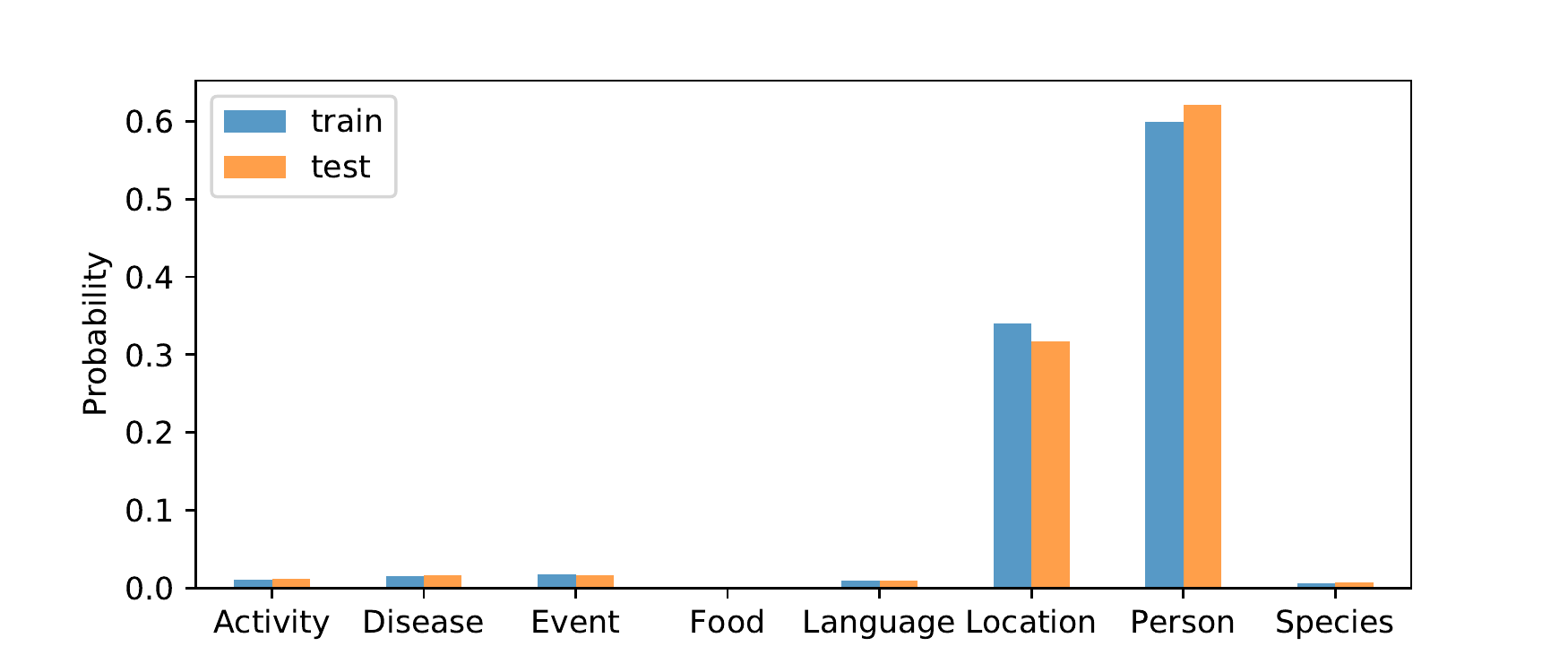}
        \caption{Types of entities.}
        \label{fig:en_type}
    \end{subfigure}
    \begin{subfigure}[b]{0.36\textwidth}
        \includegraphics[width=\textwidth]{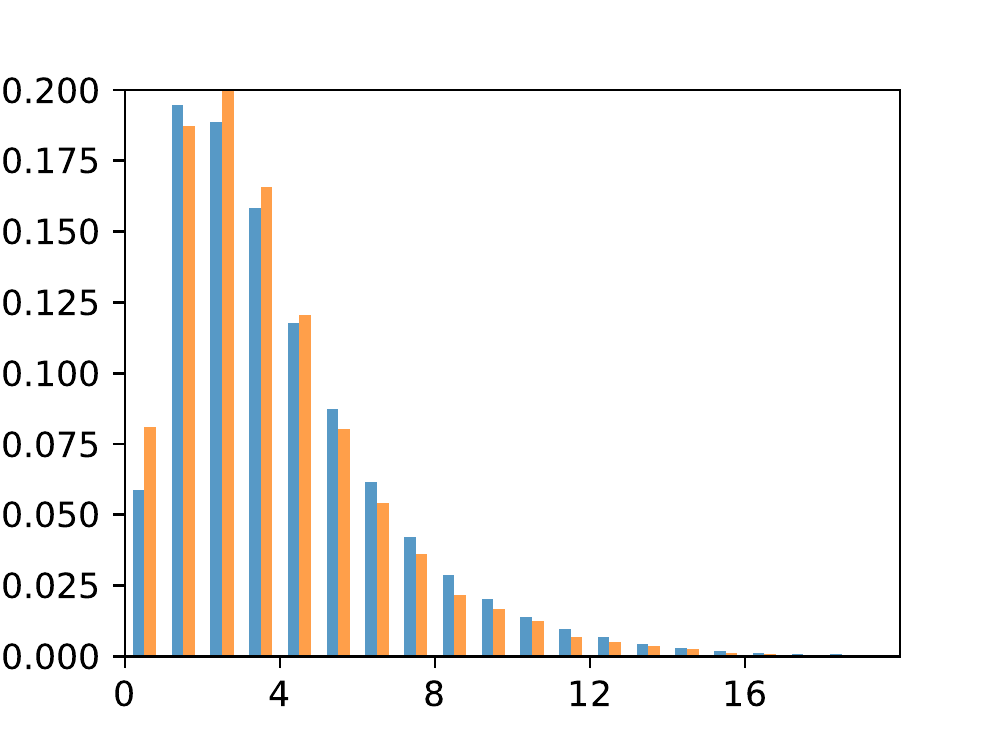}
        \caption{Number of entities per sentence.}
        \label{fig:en_num}
    \end{subfigure}
    \caption{Distribution of the entity types, and the number of entities per sentence, in the CLEF CheckThat!\ 2019 dataset. Entities are detected using DBpedia Spotlight. Note that we omit the figures for 2020 dataset, since we observe similar distributions.   }
    \label{fig:distribution}
\end{figure}

Next, as our approach makes use of entity occurrences, in Figure~\ref{fig:en_type} we show the proportion of each entity type appearing in the 2019 dataset\footnote{Similar distributions were observed for the 2020 dataset, and hence are omitted.}. In particular, it can be seen that the \textit{Person} and \textit{Location} types are the most commonly identified in the dataset, and together they account for ~90\% of all the entities detected. Figure~\ref{fig:en_num} shows the number of entities appearing in each sentence. We observe that sentences with 0-2 entities account for more than 40\% of the sentences, while sentences with 3 entities account for $\sim$15\% of sentences. The observation of the distributions of the number of entities present in each sentence further strengthens the reasons for using entity pairs (described in Section~\ref{sec:ent}).

\begin{table}[tb]
\footnotesize
\centering
\caption{A debate transcript from the CLEF'2019 CheckThat!\ dataset. Sentences are labelled check-worthy (1) or not (0). }
\label{tab:example}
\begin{tabular}{l|O|c}
Speaker & Sentence & Label \\ \hline\hline
Cruz & You know, in the past couple of weeks the Wall Street Journal had a very interesting article about the state of Arizona.  & 0  \\\hline
Cruz & Arizona put in very tough laws on illegal immigration, and the result was illegal immigrants fled the state, and what's happened there -- it was a very interesting article.  & 1  \\\hline
Cruz & Some of the business owners complained that the wages they had to pay workers went up, and from their perspective that was a bad thing.   & 0  \\\hline
Cruz & But, what the state of Arizona has seen is the dollars they're spending on welfare, on prisons, and education, all of those have dropped by hundreds of millions of dollars. & 1  \\\hline
\end{tabular}
\end{table}

\begin{table}[tb]
\footnotesize
\caption{Statistics  of  the  CLEF’2019 \& 2020  CheckThat!\ datasets.}
\label{tab:data}\centering
\begin{tabular}{l|l|r|r}
&& Training  & Testing  \\ \hline
&\# of debates/speeches & 19 & 7 \\
2019&\# of total sentences & 16,421 & 7,079 \\ 
&\# of check-worthy sentences & 433 & 110 \\ 
&\% of check-worthy sentences & 2.637\% & 2.554\% \\\hline
&\# of debates/speeches & 50 & 20 \\
2020&\# of total sentences & 42,776 & 21,514 \\ 
&\# of check-worthy sentences & 487 & 136 \\ 
&\% of check-worthy sentences & 1.138\% & 0.632\% \\\hline
\end{tabular}
\end{table}

\subsection{Models and Baselines}

In this section, we describe the tools and methods we use in our experiments, along with the baseline approaches. 
\newline

\noindent\textbf{Pre-processing:} To allow a fair comparison with previous methods, we follow the pre-processing procedure of Su et al.~\citep{su2019entity}, which includes first person resolution and co-reference resolution\footnote{the co-referent resolution considers two sentences at a time, align with that of ~\citep{su2019entity}.}. In doing so, we aim to ensure that any implied entities in the text are therefore explicitly available for analysis by the later stages of our framework (e.g.\ the language models and entity linking). In particular, we replace the first person pronouns with the speaker's name, and use the coreference resolution package\footnote{\url{https://github.com/kentonl/e2e-coref}} implemented by Lee et al.~\citep{lee2018higher}. 

\noindent \textbf{Named entity linking:} To explicitly address the entities that occur in each sentence, we deploy a named entity linking method to extract entities from each sentence. In our experiments, we use DBpedia Spotlight\footnote{\url{https://www.dbpedia-spotlight.org/}} to extract entities from each pre-processed sentence, with the confidence threshold set to 0.35, following Su et al.~\citep{su2019entity}\footnote{We note that there are better performing entity linking models, however, to maintain compatibility with the previous research we only use DBpedia Spotlight as the entity linking method in this study}.

\noindent \textbf{Entity embeddings:} We use six sources of entity embeddings to represent the embedded entity pairs\footnote{We acknowledge that these entity pair representations can also be used to calculate their similarities, however our preliminary experiments shows that such methods produce less than satisfactory results, thus we omit such setup. }, as follows:
\begin{itemize}[leftmargin=*]
    \item \textbf{Wikipedia2Vec} uses the extended skip-gram methods with a link-based measure~\citep{witten2008effective} and an anchor context model to learn the embeddings of entities. We use pre-trained Wikipedia embeddings (window size=10, iteration=10, negative=15, dimensions=300)~\citep{yamada2016joint}\footnote{\url{https://wikipedia2vec.github.io/wikipedia2vec/pretrained/}}. 

    \item \textbf{TransE}~\citep{bordes2013translating} aims to embed a triplet $e = \langle e_h, r, e_t \rangle$ into the same lower dimensional space, where $\overrightarrow{e_h} + \overrightarrow{r} = \overrightarrow{e_t}$. 
    
    \item \textbf{TransR}~\citep{linlearning}is built upon TransE, where the relation embedding is projected into a separate relation space, in order to more accurately represent the rich and diverse information between entities and relations. 
    \item \textbf{RESCAL}~\citep{nickel2011three} uses a three-way tensor learning method to model the triplet of $e = \langle e_h, r, e_t \rangle$, for more flexible representation of the relationship and entities.
    \item \textbf{DISTMult}~\citep{yang2014embedding} uses a single vector to represent both entities and the relation by simplifying the bi-linear interaction between the entity and the relation, where the relation vector is represented using the diagonal matrix of the interaction. 
    \item \textbf{ComplEx}~\citep{trouillon2016complex} uses complex embeddings and the Hermitian dot product to represent the relation between two entities, and achieved a better performance than its predecessors on the entity-linking task. 
\end{itemize}

For the TransE, TransR, RESCAL, DistMult and ComplEx models, we use triplets extracted from Freebase (FB15K)~\citep{bordes2013translating} as training data. These models are trained using code provided by Zheng et al.~\citep{zheng2020dgl}\footnote{\url{https://github.com/awslabs/dgl-ke}}.

\noindent\textbf{Language representations:} We use five different text representation models to represent each sentence, as follows:
\begin{itemize}[leftmargin=*]
    \item \textbf{TF.IDF} is a commonly used BoW model to represent the text based on the word frequencies. We include TF.IDF as a baseline.
    
    \item \textbf{BiLSTM+attention} (denote as BiLSTM+att) is widely used in the literature to learn a language model from the training data. It appeared in several solutions~\citep{hansen2019neural}, which were deployed in the CLEF'2019 CheckThat!\ Lab, and demonstrated that an LSTM-based language model can effectively represent the sentences in the check-worthiness identification task. Thus, in this paper, we use BiLSTM+att as a non-pretrained language model baseline, in order to obtain a fair comparison among all the language representation methods.

    \item \textbf{BERT}~\citep{devlin2018bert} is a state-of-the-art pre-trained language model that has been shown to be effective in many information retrieval and natural language processing tasks~\citep{macavaney2019ContextualWR,su2019ensembles,yilmaz2019applying}. 
    In this study, we are also interested in determining if the BERT model also performs well on the very specific task of check-worthy sentence identification, or if it can be enhanced by supplementary information such as entity embeddings (as discussed in the next section).

    \item \textbf{ALBERT}~\citep{lan2019albert} is a derivative of the original BERT model that. aims to reduce the number of parameters. Specifically, ALBERT uses a factorised embedding parameterisation method to decompose the vocabulary size and the hidden layer size, by projecting the vocabulary twice rather than once. Moreover, cross-layer parameter sharing and inter-sentence coherence loss are used to further reduce the need of parameters updating. ALBERT achieved a new SOTA performance with less parameters and shorter training time, compared to the original BERT model. 
    \item \textbf{RoBERTa}~\citep{liu2019roberta} aims to improve over BERT by training the model for more iterations, using longer sentence sequences, with bigger batches over more data. RoBERTa also removes the next sentence prediction objective in training. Similar to the ALBERT model, RoBERTa results in improved  performance over the standard BERT model.
\end{itemize}

We use the HuggingFace language model implementations~\citep{Wolf2019HuggingFacesTS}\footnote{\url{https://github.com/huggingface/transformers}}. Specifically, we use the BERT-Cased English model (12-layer, 768-hidden, 12-heads, 110M parameters);
the Albert-base-v2 English model (12-layer, 128-hidden, 12-heads, 1M parameters);
and the RoBERTa-base English model (12-layer, 768-hidden, 12-heads, 125M parameters). 
We fine-tune all the BERT-related language models on the training datasets. All other parameters remain at their recommended settings.

\noindent \textbf{Baselines:} We compare our generated Entity-Assisted models to the following baselines: 
\begin{itemize}[leftmargin=*]
    \item {\bf TF.IDF + SVM} (denoted by SVM(TF.IDF): We apply a SVM text classifier using TF.IDF feature vectors. We select our hyperparameters by applying cross validation on the training data. Specifically, we use the sci-kit learn SVM implementation with an RBF kernel, a C penalty of 10, and a $\gamma$ of 0.1 in our trained SVM classifier. We use class weights based on the training data to prevent the imbalanced data from compromising our experimental results. For the classification task we obtain the predicted class label for each sentence from $f^{cls}()$ (as per Equation~\eqref{eqn:cls}), while for the ranking task we obtain a score for each sentence in the range (0, 1) from $f^{rank}()$ (as per Equation~\eqref{eqn:rank}). We use the same functions for SVM(TF.IDF) \textbf{similarity} (introduced below). 
    \item {\bf SVM(TF.IDF) + Entity Similarity and Relatedness (denoted by \textbf{similarity}):} Following Su et al.~\citep{su2019entity}, we append two graph-based entity similarity and relatedness scores -- obtained using Sematch~\citep{zhu2015sematch} -- as well as a feature denoting the number of entities, to the TF.IDF feature vectors of the SVM model.
    \item {\bf BiLSTM + Att: } We deploy a BiLSTM + Att model (100 hidden units) with an attention mechanism, implemented using Tensorflow. We initialise the embedding layer of BiLSTM using the pre-trained GloVe embeddings (300 dimensions). 
    \item {\bf CLEF'2019 \& 2020 CheckThat!\ Lab leaderboards:} For the ranking task, we additionally compare with the runs of the top three groups on the official CLEF'2019 \& 2020 leader boards\footnote{From \url{https://github.com/apepa/clef2019-factchecking-task1} for 2019 results, and \url{https://github.com/sshaar/clef2020-factchecking-task5} for 2020 results.}. 
\end{itemize}

\subsection{Evaluation Metrics} 
\looseness -1 For evaluating the classification accuracy, we use the standard classification metrics (Precision, Recall, F1). Significant differences are measured using the McNemar's test.\footnote{We evaluate the classification task with only the CLEF'2019 CheckThat!\ Lab dataset, as our prior results found that it is not possible to derive meaningful results from the 2020 dataset, due to the small number of positive data in the test set.} On the other hand, for evaluating the ranking effectiveness, we apply the ranking metrics used by the CheckThat!\ Lab organisers, namely Mean Average Precision (MAP), Reciprocal Rank (RR), and Precision at rank $k$ (P@$k$, $k$=\{1,5,10,20,50\}). Means are calculated over the seven and twenty debates and speeches in the CheckThat!\ 2019 and 2020 test sets, respectively - therefore, due to the small number of rankings being evaluated, significance testing is not meaningful. Finally, note that it is not possible to evaluate the CLEF'2019 \& 2020 CheckThat!\ Lab participants' approaches using the classification metrics -- this is because the participants' runs have scores rather than predicted labels, and do not contain predictions for all sentences in the dataset. Further, it is also not possible to combine our approach with the participants' runs, since we do not have the predicted scores of the participants' runs on the training sets.

\begin{table}
\centering
\footnotesize
\caption{Classification performances on CheckThat!\ 2019 dataset, alternating language models $l_{rep}$ only. \textbf{Bold} indicate the best performance; Numbers in the Significance column indicate that the model is significantly better than the numbered model (McNemar's Test, $p$$<$$0.01$). }
\label{tab:classification-RQ1}
\begin{tabular}{|r|l|c|c|c|l|} \hline
\# & $l_{rep}$ & P & R & F1 & Significance  \\ \hline
1 & ~Random Classifier~~ & 0.01 & 0.01 & 0.01 & ~-~ \\
2 & ~SVM(TF.IDF)~~~~~~~~ & 0.01 & 0.01 & 0.01 & ~-~ \\
3 & ~BiLSTM+att~~~~~~~~~ & 0.12 & 0.07 & 0.09 & 1,2 \\
4 & ~BERT~~~~~~~~~~~~~~~ & 0.12 & 0.09 & 0.10 & 1-3 \\
5 & ~ALBERT~~~~~~~~~~~~  & \textbf{0.14} & \textbf{0.11} & \textbf{0.12} & 1-4 \\
6 & ~RoBERTa~~~~~~~~~~~  & \textbf{0.14} & \textbf{0.11} & 0.11 & 1-4 \\ \hline
\end{tabular}
\end{table}

\begin{table}
\centering
\footnotesize
\caption{Ranking performances on CheckThat!\ 2019 and 2020 dataset, alternating language models $l_{rep}$ only. \textbf{Bold} indicate the best performance in each group. }
\label{tab:ranking-RQ1}
\begin{tabular}{|r|l|c|c|c|c|c|c|c|} \hline
\# & $l_{rep}$ & MAP & MRR & P@1 & P@5 & P@10 & P@20 & P@50  \\ \hline
\multicolumn{9}{|c|}{CLEF'2019 CheckThat!\ Experimental results}  \\ \hline
1 & SVM(TF.IDF)~~~              & 0.1193 & 0.3513 & 0.1429  & \textbf{0.2571} & \textbf{0.1571} & 0.1714 & 0.1086 \\
2 & BiLSTM+att~~~~              & \textbf{0.1453} & 0.2432 & 0.1429  & 0.1429 & 0.1429 & \textbf{0.1857} & \textbf{0.1343} \\
3 & BERT~~~~~~~~~               & 0.0715 & 0.2257 & 0.1429  & 0.2000 & 0.1286 & 0.0857 & 0.0600 \\
4 & ALBERT~                     & 0.1332 & \textbf{0.4176} & \textbf{0.3098}  & 0.2000 & 0.1429 & 0.1286 & 0.0929 \\
5 & RoBERTa~                    & 0.1011 & 0.3158 & 0.2286  & 0.2000 & 0.1429 & 0.1286 & 0.0929 \\ \hline
\multicolumn{9}{|c|}{CLEF'2019 CheckThat!\ Submitted Runs}  \\ \hline
6 & Copenhagen-primary~~        & 0.1660 & 0.4176 & \textbf{0.2857}  & \textbf{0.2571} & 0.2286 & 0.1571 & 0.1229\\
7 & Copenhagen-contr.-1~~       & 0.1496 & 0.3098 & 0.1429  & 0.2000 & 0.2000 & 0.1429 & 0.1143\\
8 & Copenhagen-contr.-2~        & 0.1580 & 0.2740 & 0.1429  & 0.2286 & \textbf{0.2429} & 0.1786 & 0.1200\\
9 & TheEarthIsFlat-primary~     & 0.1597 & 0.1953 & 0.0000  & 0.2286 & 0.2143 & 0.1857 & \textbf{0.1457}\\
10 & TheEarthIsFlat-contr.-1~   & 0.1453 & 0.3158 & \textbf{0.2857}  & 0.1429 & 0.1429 & 0.1357 & 0.1171\\
11 & TheEarthIsFlat-contr.-2~   & \textbf{0.1821} & \textbf{0.4187} & \textbf{0.2857}  & 0.2286 & 0.2286 & \textbf{0.2143} & 0.1400\\
12 & IPIPAN-primary~            & 0.1332 & 0.2865 & 0.1429  & 0.1430 & 0.1715 & 0.1500 & 0.1171\\ \hline \hline
\multicolumn{9}{|c|}{CLEF'2020 CheckThat!\ Experimental results}  \\ \hline
13 & SVM(TF.IDF)~~~~~~~         & \textbf{0.0946} & 0.1531 & 0.0000  & \textbf{0.0600} & 0.0400 & \textbf{0.0450} & ~0.0240~~  \\
14 & BiLSTM+att~~~~~~~~         & 0.0151 & 0.0320 & 0.0000  & 0.0100 & 0.0150 & 0.0075 & ~0.0090~~  \\
15 & BERT~~~~~~~~~~~~~~         & 0.0262 & 0.0819 & 0.0500  & 0.0300 & 0.0250 & 0.0125 & ~0.0110     \\
16 & ALBERT~~~~~~~~~~~~         & 0.0537 & \textbf{0.2145} & \textbf{0.2000}  & 0.0800 & \textbf{0.0500} & 0.0250 & ~\textbf{0.1600}     \\
17 & RoBERTa~~~~~~~~~~~         & 0.0424 & 0.1315 & 0.1000  & 0.0600 & 0.0400 & 0.0200 & ~0.1400   \\ \hline
\multicolumn{9}{|c|}{CLEF'2020 CheckThat!\ Submitted Runs}  \\ \hline
18 & ~NLP\_IR@UNED-primary~~~~~ & \textbf{0.0867} & \textbf{0.2770} & \textbf{0.1500}  & \textbf{0.1300} & \textbf{0.0950} & \textbf{0.0725} & \textbf{0.0390}  \\
19 & ~NLP\_IR@UNED-contr.-1~    & 0.0849 & 0.2590 & \textbf{0.1500}  & 0.1200 & 0.0900 & 0.0675 & 0.0370  \\
20 & ~NLP\_IR@UNED-contr.-2~~   & 0.0408 & 0.1170 & 0.0500  & 0.0700 & 0.0450 & 0.0275 & 0.0180  \\
21 & ~UAICS-primary~~~~~~       & 0.0515 & 0.2247 & \textbf{0.1500}  & 0.0700 & 0.0500 & 0.0375 & 0.0270  \\
22 & ~UAICS-contr.-1~~~         & 0.0431 & 0.1735 & 0.1000  & 0.0500 & 0.0550 & 0.0450 & 0.0250  \\
23 & ~UAICS-contr.-2~~~         & 0.0328 & 0.1138 & 0.0500  & 0.0300 & 0.0350 & 0.0175 & 0.0190  \\
24 & ~TobbEtuP-primary~~~       & 0.0183 & 0.0326 & 0.0000  & 0.0200 & 0.0100 & 0.0100 & 0.0060  \\
25 & ~TobbEtuP-contr.-1~~~      & 0.0417 & 0.0784 & 0.0500  & 0.0300 & 0.0150 & 0.0150 & 0.0180 \\ \hline
\end{tabular}
\end{table}
\section{Experimental Results} 
\label{sec:res}
In this section, we present the results of the experiments that address RQs 1 - 4. In particular, for both the check-worthy sentence classification and ranking tasks, Sections~\ref{sec:5.1} - \ref{sec:5.4} respectively address: the effectiveness of the BERT-related language models; the usefulness of entity embeddings; the most effective combination method for representing entity pairs; and the most effective KG embedding model from which to obtain the entity embeddings. Tables~\ref{tab:classification-RQ1}, \ref{tab:classification-RQ2}, and \ref{tab:classification-RQ4} presents the attained classification results obtained on the CLEF CheckThat!\ 2019 dataset, to address RQ 1, RQs 2-3, and RQ 4 respectively. Similarly, Tables~\ref{tab:ranking-RQ1}, \ref{tab:ranking-RQ2}, and \ref{tab:ranking-RQ4} present the attained ranking performances on both the CLEF CheckThat!\ 2019 \& 2020 datasets, for RQ 1, RQs 2-3, and RQ 4 respectively. Furthermore, Tables~\ref{tab:classification-summary} \& \ref{tab:ranking-summary} summarise the performance of a salient subset of approaches on the classification and ranking tasks, respectively. 
In order to obtain further insights of this study, we conduct failure analysis and case studies in Section~\ref{sec:5.5}. Finally, in Section~\ref{sec:5.6} we summarise and discuss our main findings.

\subsection{RQ1: BERT-related Language Models vs.\ Baselines}
\label{sec:5.1}
\looseness -1 We firstly consider Table~\ref{tab:classification-RQ1}, which reports the attained accuracies when treating check-worthy sentence identification as a classification task, on the CLEF CheckThat!\ 2019 dataset. Firstly, in terms of F1, we note the relative weak performance of a classical SVM classifier with TF.IDF features (row 2), which performs equivalently to a random classifier. Indeed, while the SVM classifier has been trained using class weights to alleviate the issue of class imbalance, the low performance of SVM illustrates the difficulty of this task, and underlines that simply matching on {\em what is being said} by the speakers is insufficient to attain high accuracies on this task. Next, the BiLSTM+att classifier (row 3) markedly outperforms the random classifier, demonstrating that the deployment of pre-trained (i.e., GloVe) word embeddings allows a more flexible classifier not tied to the exact matching of tokens. Moreover, the use of the attention mechanism in BiLSTM also emphasises the importance of the context of each word. Finally, the state-of-the-art BERT-related models (BERT model, row 4; ALBERT model, row 5, and RoBERTa model, row 6) significantly outperform the random classifier, the SVM classifiers,  and the BiLSTM+att classifiers. Thus, we conclude that, when treating the task as a classification task, all of the BERT-related language models can significantly outperform the SVM and BiLSTM+att classifiers. Among all the BERT-related models, ALBERT exhibits the highest performance.

\looseness -1 Moving next to the ranking task on the 2019 dataset, Table~\ref{tab:ranking-RQ1} shows that the BERT model (row 3) performance is less than that of a classical SVM classifier using TF.IDF features (row 1). Both ALBERT (row 4) and  RoBERTa (row 5) outperform SVM(TF.IDF) and BiLSTM+att (rows 1, 2) in terms of MRR. However, in terms of MAP, both ALBERT and RoBERTa only outperform SVM(TF.IDF) (row 1), and still underperform compared to BiLSTM+att (row 2). Next, when considering the results of the ranking task on the 2020 dataset, BERT (row 15), ALBERT (row 16) and RoBERTa (row 17) models all outperform BiLSTM+att (row 14) and SVM(TF.IDF) (row 13) on both MAP and MRR. 

While the contrast between the F1 classification and the ranking results on the 2019 dataset is notable, the low classification recall for all models suggests that BERT, ALBERT, and RoBERTa (c.f.\ rows 4, 5, 6 in Table~\ref{tab:classification-RQ1}) cannot retrieve the most difficult check-worthy sentences, and hence also exhibit low MAP performances in the ranking task. From Table~\ref{tab:ranking-summary} we observe the inconsistent performances for the same language model across the 2019 and 2020 ranking datasets (i.e., row 1 vs. 16, row 4 vs. 18, row 7 vs. 20, row 10 vs. 22, row 13 vs. 24), we postulate that this may be caused by the markedly different proportion of positive examples in the two test sets (as illustrated by the percentage of the check-worthy sentences in Table~\ref{tab:data}).
Overall, in answer to RQ1, we conclude that while the BERT, ALBERT, and RoBERTa models perform well at classifying check-worthy sentences, for ranking they are most effective at higher rank sentences. On both tasks, ALBERT performs the best among the BERT-related language models. 

\begin{table}
\centering
\footnotesize
\caption{Classification performances on CheckThat!\ 2019 dataset, alternating language models $l_{rep}$ and entity embedding models $m_{ent}$, and entity representation combination models $e_{com}$. \textbf{Bold} indicate the best performance; Numbers in the Significance column indicate that the model is significantly better than the numbered model (McNemar's Test, $p$$<$$0.01$). }
\begin{tabular}{|r|l|l|l|c|c|c|l|} \hline
\# & $l_{rep}$ & $m_{ent}$ & $e_{com}$ & P & R & F1 & Significance  \\ \hline
1   & SVM(TF.IDF)   & -                 & -             & 0.01      & 0.01      & 0.01      & -         \\
2   & SVM(TF.IDF)   & Wikipedia2Vec     & similarity    & 0.04      & 0.03      & 0.03      & 1             \\
3   & SVM(TF.IDF)   & Wikipedia2Vec     & emb\_concat   & 0.06      & 0.05      & 0.05      & 1,2           \\
4   & SVM(TF.IDF)   & Wikipedia2Vec     & emb\_prod     & 0.05      & 0.04      & 0.04      & 1,2           \\\hline
5   & BiLSTM+att    & -                 & -             & 0.12      & 0.07      & 0.09      & 1-4           \\
6   & BiLSTM+att    & Wikipedia2Vec     & similarity    & 0.12      & 0.08      & 0.1       & 1-5           \\
7   & BiLSTM+att    & Wikipedia2Vec     & emb\_concat   & 0.13      & 0.1       & 0.11      & 1-6           \\
8   & BiLSTM+att    & Wikipedia2Vec     & emb\_prod     & 0.12      & 0.09      & 0.1       & 1-5           \\\hline
9   & BERT          & -                 & -             & 0.12      & 0.09      & 0.1       & 1-5           \\
10  & BERT          & Wikipedia2Vec     & similarity    & 0.12      & 0.1       & 0.11      & 1-6           \\
11  & BERT          & Wikipedia2Vec     & emb\_concat   & 0.19      & 0.11      & 0.14      & 1-10, 13      \\
12  & BERT          & Wikipedia2Vec     & emb\_prod     & 0.18      & 0.11      & 0.13      & 1-10, 13      \\\hline
13  & ALBERT        & -                 & -             & 0.14      & 0.11      & 0.12      & 1-10          \\
14  & ALBERT        & Wikipedia2Vec     & similarity    & 0.14      & 0.14      & 0.14      & 1-10, 13      \\
15  & ALBERT        & Wikipedia2Vec     & emb\_concat   & \textbf{0.22}      & \textbf{0.15}      & \textbf{0.18}      & 1-14, 17-20   \\
16  & ALBERT        & Wikipedia2Vec     & emb\_prod     & 0.20      & 0.14      & 0.16      & 1-14, 17, 18  \\\hline
17  & RoBERTa       & -                 & -             & 0.14      & 0.11      & 0.12      & 1-10          \\
18  & RoBERTa       & Wikipedia2Vec     & similarity    & 0.14      & 0.13      & 0.13      & 1-10          \\
19  & RoBERTa       & Wikipedia2Vec     & emb\_concat   & 0.21      & \textbf{0.15}      & 0.17      & 1-14, 17, 18  \\ 
20  & RoBERTa       & Wikipedia2Vec     & emb\_prod     & 0.19      & 0.14      & 0.16      & 1-14, 17, 18  \\\hline
\end{tabular}
\label{tab:classification-RQ2}
\end{table}

\begin{table}
\centering
\footnotesize
\caption{Ranking performances on CheckThat!\ 2019 dataset, alternating language models $l_{rep}$ and entity embedding models $m_{ent}$, and entity representation combination models $e_{com}$. \textbf{Bold} indicate the best performance. }
\begin{tabular}{|r|l|l|l|c|c|c|c|c|c|c|} \hline
\# & $l_{rep}$ & $m_{ent}$ & $e_{com}$ & MAP & MRR & P@1 & P@5 & P@10 & P@20 & P@50  \\ \hline
\multicolumn{11}{|c|}{CLEF'2019 CheckThat!\ Experimental results}  \\ \hline
1  & SVM(TF.IDF)~~~ & -               & -           & 0.1193 & 0.3513 & 0.1429 & 0.2571 & 0.1571 & 0.1714 & 0.1086  \\
2  & SVM(TF.IDF)~~~ & Wikepedia2Vec~~ & similarity~ & 0.1263 & 0.3254 & 0.2857 & 0.2000 & 0.2000 & 0.1286 & 0.0915  \\
3  & SVM(TF.IDF)~~~ & Wikepedia2Vec~~ & emb\_concat & 0.1332 & 0.3361 & 0.3254 & 0.2000 & 0.2000 & 0.1286 & 0.0915  \\ 
4  & SVM(TF.IDF)~~~ & Wikepedia2Vec~~ & emb\_prod   & 0.1332 & 0.3361 & 0.3254 & 0.2000 & 0.2000 & 0.1286 & 0.0915  \\\hline
5  & BiLSTM+att~~~~ & -               & -           & 0.1453 & 0.2432 & 0.1429 & 0.1429 & 0.1429 & 0.1857 & 0.1343  \\
6  & BiLSTM+att~~~~ & Wikepedia2Vec~~ & similarity~ & 0.0715 & 0.2857 & 0.2432 & 0.1429 & 0.1286 & 0.0714 & 0.0314  \\
7  & BiLSTM+att~~~~ & Wikepedia2Vec~~ & emb\_concat & 0.0659 & 0.3361 & 0.2857 & 0.1429 & 0.1429 & 0.0714 & 0.0314  \\ 
8  & BiLSTM+att~~~~ & Wikepedia2Vec~~ & emb\_prod   & 0.0659 & 0.3158 & 0.2000 & 0.1429 & 0.1286 & 0.0714 & 0.0714  \\\hline
9  & BERT~~~~~~~~~  & -               & -           & 0.0715 & 0.2257 & 0.1429 & 0.2000 & 0.1286 & 0.0857 & 0.0600  \\
10 & BERT~~~~~~~~~  & Wikepedia2Vec~~ & similarity~ & 0.0826 & 0.3158 & 0.3098 & 0.2000 & 0.1286 & 0.0929 & 0.0600  \\
11 & BERT~~~~~~~~~  & Wikepedia2Vec~~ & emb\_concat & 0.1011 & \textbf{0.6196} & \textbf{0.3361} & 0.1714 & 0.1429 & 0.0929 & 0.0686  \\ 
12 & BERT~~~~~~~~~  & Wikepedia2Vec~~ & emb\_prod   & 0.0826 & 0.3361 & \textbf{0.3361} & 0.1429 & 0.1429 & 0.0929 & 0.0929  \\\hline
13 & ALBERT~        & -               & -           & 0.1332 & 0.4176 & 0.3098 & 0.2000 & 0.1429 & 0.1286 & 0.0929  \\
14 & ALBERT~        & Wikepedia2Vec~~ & similarity~ & 0.1453 & 0.4176 & \textbf{0.3361} & 0.2286 & 0.2000 & 0.1286 & 0.1286  \\
15 & ALBERT~        & Wikepedia2Vec~~ & emb\_concat & \textbf{0.1580} & \textbf{0.6196} & 0.3098 & 0.2857 & \textbf{0.2571} & \textbf{0.2286} & \textbf{0.2286}  \\ 
16 & ALBERT~        & Wikepedia2Vec~~ & emb\_prod   & 0.1332 & 0.4187 & \textbf{0.3361} & 0.2571 & \textbf{0.2571} & 0.2000 & 0.1286  \\\hline
17 & RoBERTa~       & -               & -           & 0.1011 & 0.3158 & 0.2286 & 0.2000 & 0.1429 & 0.1286 & 0.0929  \\
18 & RoBERTa~       & Wikepedia2Vec~~ & similarity  & 0.1263 & 0.4176 & \textbf{0.3361} & 0.2286 & 0.2000 & 0.1286 & 0.0929  \\
19 & RoBERTa~       & Wikepedia2Vec~~ & emb\_concat & 0.1453 & 0.4176 & \textbf{0.3361} & \textbf{0.2857} & \textbf{0.2571} & 0.2000 & \textbf{0.2286} \\ 
20 & RoBERTa~       & Wikepedia2Vec~~ & emb\_prod   & 0.1332 & 0.4187 & \textbf{0.3361} & 0.2571 & 0.2000 & 0.2000 & 0.1286 \\ \hline
\end{tabular}
\label{tab:ranking-RQ2}
\end{table}

\subsection{RQ2: Using Entity Embeddings}
\label{sec:5.2}
We now examine the impact of entities at improving the classification accuracy. Firstly, from Table~\ref{tab:classification-RQ2}, we note that the F1 performance of the SVM classifier is improved by adding the entity similarity scores (row 2 vs row 1), echoing our earlier observations~\citep{su2019entity} for ranking, on the 2019 dataset. Similarly, adding entity emb\_prod (row 4) element-wise product operation between language representation and entity pair representation and entity emb\_concat (row 3) concatenating language representation and entity pair representation also improve the SVM classifier's performance, using the 2019 dataset, in terms of precision, recall and F1 compared to SVM(TF.IDF) without entity information. Next, we observe that all of the neural language models (i.e., BiLSTM+att, BERT, ALBERT, RoBERTa) also exhibit a significantly improved accuracy when combined with entity embeddings (rows 7 \& 8 vs.\ 5; rows 11 \& 12 vs.\ 9; rows 15 \& 16 vs.\ 13; rows 19 \& 20 vs\ 17). On the contrary, even though we do observe an improvement when the neural models are combined with entity similarities (row 6 vs.\ 5; row 10 vs.\ 9, row 14 vs.\ 13; row 18 vs.\ 17), the improvement is not significant. Moreover, combining the neural models with the entity embedding information, both entity combination methods significantly outperform the corresponding language model when combined with simpler entity similarities. Indeed, our proposed entity-assisted ALBERT classifiers using the emb\_concat method (row 16) attains the highest overall classification performance (an F1 score of 0.18).
Table~\ref{tab:classification-RQ2} further shows that almost all neural models with all types of entity embeddings outperform the corresponding language models alone, in terms of F1. 
Thus, we conclude that the entity information (the entity similarity and embedding for the SVM classifier, the entity embeddings in the neural language models) can indeed improve the classification accuracy in the identification of check-worthy sentences. 

\looseness -1 Turning to the ranking task, in Table~\ref{tab:ranking-RQ2}, we observe that the use of entities (i.e., entity similarities, emb\_concat, and emb\_prod) enhances most of the approaches: the effectiveness of the SVM classifier is enhanced on MAP, P@1, and P@10. On the other hand, while BiLSTM+att is enhanced for MRR and P@1, when combined with any type of entity information, the MAP performances are damaged by the entity embeddings (rows 6-8 vs.\ 5). Finally, the BERT-related models (i.e., BERT, ALBERT, RoBERTa) are enhanced by all three types of entity information, regardless of the entity embedding combination model used, in terms of MAP, MRR, and P@1  (rows 10-12 vs.\ 9; rows 14-16 vs.\ 13; rows 18-20 vs.\ 17). 
When tested on the 2020 dataset, Table~\ref{tab:ranking-summary} shows that the ComplEx KG embedding model together with the emb\_concat method, consistently improves all the neural networks' performance on all metrics. Thus, we conclude that entity embeddings can consistently enhance the BiLSTM+att models for ranking on high precision metrics such as MRR and P@1, as well as enhance the SVM(TF.IDF) and neural language models (i.e., BERT, ALBERT, RoBERTa) across the evaluation metrics.

Therefore, in response to RQ2, we conclude that using entity embeddings -- regardless of the KG embedding model from which the entity embeddings are obtained -- does help to improve the BERT-related language models' performance, on both precision and recall for the classification task, and on MAP, MRR and P@1 for the ranking tasks.

\subsection{RQ3: Entity Representation}
\label{sec:5.3}

\looseness -1 When considering identifying check-worthy sentences as a classification task, Table~\ref{tab:classification-RQ2} shows that all of the SVM(TF.IDF) and BERT-related language models are significantly improved when combined with entity embeddings, over the language models alone or with entity similarities. 
Meanwhile, we observe that using emb\_concat only marginally outperforms emb\_prod, without significant differences (row 3 vs.\ 4; row 7 vs.\ 8; row 11 vs.\ 12; row 15 vs.\ 16; row 19 vs.\ 20). 
Moreover, the ALBERT model with the emb\_concat method using the Wikipedia2Vec KG embedding model (row 15) achieves the highest F1 score among all of the tested models shown in Table~\\ref{tab:classification-RQ2}. Thus, we conclude that using entity embeddings is more effective than using the entity similarity method suggested by Su et al.~\citep{su2019entity} on the classification task, which uses graph distance for estimating the similarity between entities.

Next, when considering the ranking task, Table~\ref{tab:ranking-RQ2} shows that the BiLSTM+att, and BERT-related languages models all exhibit improved MRR and P@1 when combined with entity embeddings using the concatenation method, outperforming the entity similarity method (rows 3 \& 4 vs.\ 2; 7 \& 8 vs.\ 6; 11 \& 12 vs.\ 10; rows 15 \& 16 vs. 14; rows 19 \& 20 vs. row 18). In terms of the entity representation methods for the embedded entities, emb\_concat and emb\_prod perform similarly for SVM and BiLST+att (rows 3 \& 4, rows 7 \& 8), however for the BERT models emb\_concat exhibits an 84\% increase over emb\_prod (row 11 vs.\ 12). When combining the embedded entities with ALBERT and RoBERTa, we also observe that emb\_concat consistently exhibits a performance increase over emb\_prod (row 15 vs.\ 16; row 19 vs.\ 20). Thus, we conclude that for the ranking task, the emb\_concat model is more effective than emb-prob, and both embedding methods are more effective than the entity similarity baseline (rows 2, 6, 10, 14, 18). 

Overall, in answer to RQ3, we conclude that using embedding entities obtained from KG embedding models, regardless of the representation method, improves all three BERT-based language representations better than the entity similarity information, with emb\_concat exhibiting the highest effectiveness on both for the classification task (using the 2019 dataset) and the ranking task (using the 2019 \& 2020 datasets).

\begin{table}
\centering
\footnotesize
\caption{Classification performances on CheckThat!\ 2019 dataset, using emb\_concat as entity representation combination method, while alternating language models $l_{rep}$ and entity embedding models $m_{ent}$. \textbf{Bold} indicate the best performance; Numbers in the Significance column indicate that the model is significantly better than the numbered model (McNemar's Test, $p$$<$$0.01$). }
\begin{tabular}{|r|l|l|c|c|c|l|} \hline
\# & $l_{rep}$ & $m_{ent}$ & P & R & F1 & Significance  \\ \hline
1  & ~SVM(TF.IDF)~~~~~~~~ & ~Wikipedia2Vec~ & 0.06 & 0.05 & 0.05 & ~-~          \\
2  & ~SVM(TF.IDF)~~~~~~~~ & ~TransE~~~~~~~~ & 0.06 & 0.05 & 0.05 & -~           \\
3  & ~SVM(TF.IDF)~~~~~~~~ & ~TransR~~~~~~~~ & 0.06 & 0.05 & 0.05 & -~           \\
4  & ~SVM(TF.IDF)~~~~~~~~ & ~RESCAL~~~~~~~~ & 0.06 & 0.05 & 0.05 & -~           \\
5  & ~SVM(TF.IDF)~~~~~~~~ & ~DistMult~~~~~~ & 0.07 & 0.05 & 0.06 & -            \\
6  & ~SVM(TF.IDF)~~~~~~~~ & ~ComplEx~~~~~~~ & 0.07 & 0.05 & 0.06 & -            \\
7  & ~BiLSTM+att~~~~~~~~~ & ~Wikipedia2Vec~ & 0.13 & 0.10 & 0.11 & 1-6          \\
8  & ~BiLSTM+att~~~~~~~~~ & ~TransE~~~~~~~~ & 0.11 & 0.08 & 0.09 & 1-6          \\
9  & ~BiLSTM+att~~~~~~~~~ & ~TransR~~~~~~~~ & 0.12 & 0.08 & 0.09 & 1-6          \\
10 & ~BiLSTM+att~~~~~~~~~ & ~RESCAL~~~~~~~~ & 0.12 & 0.08 & 0.10 & 1-6,8,9      \\
11 & ~BiLSTM+att~~~~~~~~~ & ~DistMult~~~~~~ & 0.13 & 0.12 & 0.12 & 1-10         \\
12 & ~BiLSTM+att~~~~~~~~~ & ~ComplEx~~~~~~~ & 0.14 & 0.13 & 0.13 & 1-10         \\
13 & ~BERT~~~~~~~~~~~~~~  & ~Wikipedia2Vec~ & 0.19 & 0.11 & 0.14 & 1-11         \\
14 & ~BERT~~~~~~~~~~~~~~~ & ~TransE~~~~~~~~ & 0.19 & 0.10 & 0.13 & 1-11         \\
15 & ~BERT~~~~~~~~~~~~~~~ & ~TransR~~~~~~~~ & 0.19 & 0.11 & 0.14 & 1-11         \\
16 & ~BERT~~~~~~~~~~~~~~  & ~RESCAL~~~~~~~~ & 0.19 & 0.11 & 0.14 & 1-11         \\
17 & ~BERT~~~~~~~~~~~~~~  & ~DistMult~~~~~~ & 0.19 & 0.12 & 0.15 & 1-16         \\
18 & ~BERT~~~~~~~~~~~~~~  & ~ComplEx~~~~~~~ & 0.20 & 0.13 & 0.15 & 1-16         \\
19 & ~ALBERT~~~~~~~~~~~~  & ~Wikipedia2Vec~ & 0.22 & 0.15 & 0.18 & 1-18         \\
20 & ~ALBERT~~~~~~~~~~~~  & ~TransE~~~~~~~~ & 0.22 & 0.14 & 0.17 & 1-18         \\
21 & ~ALBERT~~~~~~~~~~~~  & ~TransR~~~~~~~~ & 0.23 & 0.14 & 0.18 & 1-18         \\
22 & ~ALBERT~~~~~~~~~~~~  & ~RESCAL~~~~~~~~ & 0.24 & 0.15 & 0.19 & 1-21, 25-28  \\
23 & ~ALBERT~~~~~~~~~~~~  & ~DistMult~~~~~~ & 0.24 & 0.15 & 0.19 & 1-21, 25-28  \\
24 & ~ALBERT~~~~~~~~~~~~  & ~ComplEx~~~~~~~ & \textbf{0.25} & \textbf{0.16} & \textbf{0.20} & 1-22, 25-30  \\
25 & ~RoBERTa~~~~~~~~~~~  & ~Wikipedia2Vec~ & 0.21 & 0.15 & 0.17 & 1-18         \\
26 & ~RoBERTa~~~~~~~~~~~  & ~TransE~~~~~~~~ & 0.21 & 0.14 & 0.16 & 1-18         \\
27 & ~RoBERTa~~~~~~~~~~~  & ~TransR~~~~~~~~ & 0.21 & 0.15 & 0.17 & 1-18         \\
28 & ~RoBERTa~~~~~~~~~~~  & ~RESCAL~~~~~~~~ & 0.20 & 0.14 & 0.16 & 1-18         \\
29 & ~RoBERTa~~~~~~~~~~~  & ~DistMult~~~~~~ & 0.23 & 0.15 & 0.18 & 1-18, 25-28  \\
30 & ~RoBERTa~~~~~~~~~~~  & ~ComplEx~~~~~~~ & 0.24 & 0.14 & 0.18 & 1-18  25-28 \\ \hline
\end{tabular}
\label{tab:classification-RQ4}
\end{table}
\vspace{0.5\baselineskip}

\begin{table}
\centering
\footnotesize
\caption{Ranking performances on CheckThat!\ 2019 dataset, using emb\_concat as entity representation combination method, while alternating language models $l_{rep}$ and entity embedding models $m_{ent}$. \textbf{Bold} indicate the best performance. }
\begin{tabular}{|r|l|l|c|c|c|c|c|c|c|} \hline
\# & $l_{rep}$ & $m_{ent}$ & MAP & MRR & P@1 & P@5 & P@10 & P@20 & P@50  \\ \hline
1  & SVM(TF.IDF)~~~ & Wikepedia2Vec~~ & 0.1332               & 0.3361               & 0.3254               & 0.2000               & 0.2000               & 0.1286               & 0.0915                \\
2  & SVM(TF.IDF)~~~ & TransE~~~~~~~~~ &     0.1332               & 0.3361               & 0.3254               & 0.2000               & 0.2000               & 0.1286               & 0.0915           \\
3  & SVM(TF.IDF)~~~ & TransR~~~~~~~~~ &   0.1263               & 0.5714               & 0.2857               & 0.1714               & 0.1429               & 0.0929               & 0.0929                \\
4  & SVM(TF.IDF)~~~ & RESCAL~~~~~~~~~ & 0.1453               & 0.4176               & \textbf{0.3361}               & 0.2857               & 0.2571               & 0.2000               & \textbf{0.2286}                \\
5  & SVM(TF.IDF)~~~ & DISTMult~~~~~~~ &     0.1453               & 0.3158               & 0.2857               & 0.2857               & 0.2000               & 0.2286               & 0.2000                \\
6  & SVM(TF.IDF)~~~ & ComplEx~~~~~~~~ &     0.1496               & 0.4187               & 0.3098               & 0.2857               & 0.2571               & 0.2000               & 0.1286                \\
7  & BiLSTM+att~~~~ & Wikepedia2Vec~~ & 0.0659               & 0.3361               & 0.2857               & 0.1429               & 0.1429               & 0.0714               & 0.0314                \\
8  & BiLSTM+att~~~~ & TransE~~~~~~~~~ & 0.0659               & 0.3158               & 0.2857               & 0.1429               & 0.1429               & 0.1429               & 0.0714                \\
9  & BiLSTM+att~~~  & TransR~~~~~~~~~ & 0.0715               & 0.3158               & 0.2432               & 0.1429               & 0.1286               & 0.0714               & 0.0314                \\
10 & BiLSTM+att~~~  & RESCAL~~~~~~~~~ & 0.0659               & 0.3361               & 0.2857               & 0.1429               & 0.1429               & 0.0714               & 0.0314                \\
11 & BiLSTM+att~~~  & DISTMult~~~~~~~ & 0.0659               & 0.3158               & 0.2000               & 0.1429               & 0.1429               & 0.1286               & 0.0714                \\
12 & BiLSTM+att~~~  & ComplEx~~~~~~~~ & 0.0715               & 0.2257               & 0.1286               & 0.1429               & 0.1429               & 0.1857               & 0.1343                \\
13 & BERT~~~~~~~~~  & Wikepedia2Vec~~ & 0.1011               & \textbf{0.6196}               & \textbf{0.3361}               & 0.1714               & 0.1429               & 0.0929               & 0.0686                \\
14 & BERT~~~~~~~~~  & TransE~~~~~~~~~ & 0.1011               & 0.5714               & 0.3098               & 0.2000               & 0.1714               & 0.1286               & 0.0929                \\
15 & BERT~~~~~~~~~  & TransR~~~~~~~~~ & 0.1011               & \textbf{0.6196}               & 0.3098               & 0.1714               & 0.0929               & 0.0929               & 0.0686                \\
16 & BERT~~~~~~~~~  & RESCAL~~~~~~~~~ & 0.1263               & 0.5714               & 0.2857               & 0.1714               & 0.1429               & 0.0929               & 0.0929                \\
17 & BERT~~~~~~~~~  & DISTMult~~~~~~~ & 0.1263               & \textbf{0.6196}               & 0.3098               & 0.2571               & 0.1429               & 0.0929               & 0.0929                \\
18 & BERT~~~~~~~~~  & ComplEx~~~~~~~~ & 0.1453               & \textbf{0.6196}               & \textbf{0.3361}               & 0.2857               & 0.1714               & 0.1286               & 0.0929                \\
19 & ALBERT~        & Wikepedia2Vec~~ & 0.1580               & \textbf{0.6196}               & 0.3098               & 0.2857               & 0.2571               & 0.2286               & \textbf{0.2286}                \\
20 & ALBERT~~~~~~~  & TransE~~~~~~~~~ & 0.1332               & 0.4176               & \textbf{0.3361}               & 0.1429               & 0.1429               & 0.1286               & 0.0929                \\
21 & ALBERT~~~~~~~  & TransR~~~~~~~~~ & 0.1263               & 0.3158               & 0.3098               & 0.2000               & 0.2286               & 0.1286               & 0.0929                \\
22 & ALBERT~~~~~~~  & RESCAL~~~~~~~~~ & 0.1332               & 0.5714               & 0.3098               & 0.2286               & 0.2000               & 0.1286               & 0.0929                \\
23 & ALBERT~~~~~~~  & DISTMult~~~~~~~ & 0.1580               & 0.4176               & 0.2857               & 0.2000               & 0.1429               & 0.1429               & 0.0929                \\
24 & ALBERT~~~~~~~  & ComplEx~~~~~~~~ & \textbf{0.1821}               & \textbf{0.6196}               & \textbf{0.3361}               & \textbf{0.3098}               & \textbf{0.2857}               & \textbf{0.2571}               & 0.1286                \\
25 & RoBERTa~       & Wikepedia2Vec~~ & 0.1453               & 0.4176               & \textbf{0.3361}               & 0.2857               & 0.2571               & 0.2000               & 0.2286                \\
26 & RoBERTa~~~~~~  & TransE~~~~~~~~~ & 0.1332               & 0.4176               & 0.2857               & 0.2571               & 0.2000               & 0.2000               & 0.1286                \\
27 & RoBERTa~~~~~~  & TransR~~~~~~~~~ & 0.1263               & 0.4176               & 0.2000               & 0.2857               & 0.2000               & 0.2286               & 0.2000                \\
28 & RoBERTa~~~~~~  & RESCAL~~~~~~~~~ & 0.1453               & 0.3158               & 0.2857               & 0.2857               & 0.2000               & 0.2286               & 0.2000                \\
29 & RoBERTa~~~~~~  & DISTMult~~~~~~~ & 0.1496               & 0.4187               & 0.3098               & 0.2857               & 0.2571               & 0.2000               & 0.1286                \\
30 & RoBERTa~~~~~~  & ComplEx~~~~~~~~ & 0.1660               & 0.5714               & \textbf{0.3361}               & \textbf{0.3098}               & 0.2000               & \textbf{0.2571}               & \textbf{0.2286}   \\ \hline            
\end{tabular}
\label{tab:ranking-RQ4}
\end{table}

\subsection{RQ4: KG Embedding Model}
\label{sec:5.4}
Finally, we consider which KG embedding model is the most effective in providing entity embeddings for identifying the check-worthy sentences. Table~\ref{tab:classification-RQ4} shows the results obtained by combining different KG entity embedding models with the various language representations for the classification task. We observe that ComplEx does not significantly outperform embedded Wikipedia2Vec when combined with SVM(TF.IDF) (row 6 vs. 1), but consistently and significantly outperforms Wikipedia2Vec, TransE, TransR and RESCAL (row 12 vs.\ rows 7-10; row 18 vs.\ 13-16; row 24 vs.\ 19-21; row 30 vs.\ 25-28) for all the neural language representation models we use. However, while ComplEx does not significantly outperform DistMult, across all language representation models, it does exhibit an average of 1\% absolute improvement in F1 over the DistMul KG embeddings (see row 12 vs.\ 11, row 18 vs.\ 17, row 24 vs.\ 23, row 30 vs.\ 29). The results are expected, given previous reported results in the literature, since ComplEx and DistMult indeed outperform other KG embedding models on KG embedding link prediction task~\cite{trouillon2016complex,yang2014embedding}.

For the ranking task, Table~\ref{tab:ranking-RQ4} shows that ALBERT + ComplEx achieves the best performance among our experiments, obtaining 0.1821, a tie with the best performing run in the official leader board, on the 2019 dataset. Moreover, it also shows that ALBERT + ComplEx obtains the highest MAP in the 2020 dataset among all the models we tested, as well as the models in the leader board. Under further investigation, we found that ALBERT + ComplEx successfully identified the single check-worthy sentence within one debate of the test set, and therefore obtained the highest improvement on MAP. 

We further conducted a case study in order to understand why ComplEx can achieve the best performance consistently. Table~\ref{tab:case_study} presents 2 cases where ComplEx model togehter with ALBERT language model successfully identified the check-worthy sentences, while other entity embedding models did not. Upon further investigation, we found that in these two cases, entity 1 and entity 2 are not related to each other closely, while Wikepedia did not mention entity 2 in the articles about entity 1. We therefore postulate that ComplEx is able to identify hidden relations between two weakly associated entities better than other entity embedding models. 

Overall, we conclude that among all 6 KG embedding models we tested, ComplEx produces consistently the highest performance.

\begin{table}
\centering
\footnotesize
\caption{Two sentences that are correctly identified as check-worthy using ALBERT, ComplEx entity embedding model, and emb\_concat model, but are otherwise not identified. }
\begin{tabular}{|l|O|l|l|} \hline
Speaker & Sentence & Entity 1 & Entity 2\\ \hline
Donald Trump  & They want to take away your good health care, and essentially use socialism to turn America into Venezuela and Democrats want to totally open the borders.                                                                            & Venezuela      & Democrat      \\\hline
Donald Trump~ & And one state said -- you know, it was interesting, one of the states we won, Wisconsin -- I didn't even realize this until fairly recently -- that was the one state Ronald Reagan didn't win when he ran the board his second time. & Wisconsin     & Ronald Reagan \\ \hline
\end{tabular}
\label{tab:case_study}
\end{table}

\begin{table}[h!]
\footnotesize
\caption{Classification performances on CheckThat!\ 2019 dataset. \textbf{Bold} indicate the best performance; Numbers in the column \textit{Significance} indicate that the model is significantly better than the numbered model (McNemar's Test, $p$$<$$0.01$). }
\centering
\begin{tabular}{|c|c|c|c|ccc|c|}
\hline
\# & $l_{rep}$          &$m_{ent}$      & $e_{com} $      & P     & R     & F1  &Significance\\\hline\hline
 1 & Random Classifier  & -             & -             & 0.01  & 0.01  & 0.01  & - \\\hline
 2 & SVM(TF.IDF)        & -             & -             & 0.01  & 0.01  & 0.01  & - \\
 3 & SVM(TF.IDF)        & Wikipedia2Vec & emb\_concat    & 0.06  & 0.05  & 0.05  & 1,2\\
 4 & SVM(TF.IDF)        & ComplEx       & emb\_concat    & 0.07  & 0.05  & 0.06  & 1-2\\ \hline

 5 & BiLSTM+att         & -             & -             & 0.12  & 0.07  & 0.09  & 1-4\\ 
 6 & BiLSTM+att         & Wikipedia2Vec & emb\_concat    & 0.13  & 0.10  & 0.11  & 1-5\\ 
 7 & BiLSTM+att         & ComplEx       & emb\_concat    & 0.14  & 0.13  & 0.13  & 1-6\\ \hline
 8 & BERT               & -             & -             & 0.12  & 0.09  & 0.10  & 1-5\\ 
 9 & BERT               & Wikipedia2Vec & emb\_concat    & 0.19  & 0.11  & 0.14  & 1-8\\ 
 10 & BERT               & ComplEx       & emb\_concat    & 0.20  & 0.13  & 0.15  & 1-9\\ \hline
 11 & ALBERT            & -             & -             & 0.14  & 0.11  & 0.12  & 1-6,8\\ 
 12 & ALBERT            & Wikipedia2Vec & emb\_concat    & 0.22  & 0.15  & 0.18  & 1-10,13\\ 
 13 & ALBERT            & ComplEx       & emb\_concat    & \textbf{0.25}  & \textbf{0.16}  & \textbf{0.20}  & 1-12,14-16\\ \hline
 14 & RoBERTa           & -             & -             & 0.14  & 0.11  & 0.11  & 1-6,8\\ 
 15 & RoBERTa           & Wikipedia2Vec & emb\_concat    & 0.21  & 0.15  & 0.17  & 1-11,14\\ 
 16 & RoBERTa           & ComplEx       & emb\_concat    & 0.24  & 0.14  & 0.18  & 1-12,14,15\\ 
 
 \hline\hline
\end{tabular}
\label{tab:classification-summary}
\end{table}

\begin{table*}[h!]
\footnotesize
\caption{Performances on the ranking metrics on both CLEF' 2019 \& 2020 CheckThat!\ dataset ). \textbf{Bold} denotes the best performance for a given measure in a given year.}
\label{tab:ranking-summary}
\centering
\begin{tabular}{|c|c|c|c|ccccccc|}
\hline
\# & $l_{rep}$&$m_{ent}$ &$e_{com}$  & MAP     & MRR     & P@1       & P@5       & P@10      & P@20      & P@50    \\
\hline
\multicolumn{11}{|c|}{Experimental Results using CLEF'2019 CheckThat!\ dataset} \\
\hline
1&SVM(TF.IDF)   &-              & -         & 0.1193  & 0.3513  & 0.1429    & 0.2571    & 0.1571    & 0.1714    & 0.1086 \\
2&SVM(TF.IDF)   &Wikepedia2Vec  &emb\_concat & 0.1332  & 0.3361  & 0.3254    & 0.2000    & 0.2000    & 0.1286    & 0.0915 \\
3&SVM(TF.IDF)   &ComplEx        &emb\_prod   & 0.1332  & 0.3158  & 0.3098    & 0.2000    & 0.2571    & 0.1429    & 0.0929 \\\hline

4&BiLSTM+att    & -             & -         & 0.1455  & 0.2432  & 0.1429    & 0.1429    & 0.1429    & 0.1857    & 0.1343 \\ 
5&BiLSTM+att    &Wikepedia2Vec  &emb\_concat & 0.0659  & 0.3361  & 0.2857    & 0.1429    & 0.1429    & 0.0714    & 0.0314 \\ 
6&BiLSTM+att   &ComplEx        &emb\_concat & 0.0715  & 0.2257  & 0.1286    & 0.1429    & 0.1429    & 0.1857    & 0.1343 \\\hline
7&BERT         & -             & -         & 0.0715  & 0.2257  & 0.1429    & 0.2000    & 0.1286    & 0.0857    & 0.0600 \\
8&BERT         &Wikepedia2Vec  &emb\_concat & 0.1011  & \textbf{0.6196}  & \textbf{0.3361}    & 0.1714    & 0.1429    & 0.0929    & 0.0686 \\ 
9&BERT         &ComplEx        &emb\_concat & 0.1011  & \textbf{0.6196}  & \textbf{0.3361}    & 0.2857    & 0.1714    & 0.1286    & 0.0929 \\\hline
10&ALBERT & -             & -         & 0.1332  & 0.4176  & 0.3098    & 0.2000    & 0.1429    & 0.1286    & 0.0929 \\
11&ALBERT &Wikepedia2Vec  &emb\_concat & 0.1580  & \textbf{0.6196}  & 0.3098    & 0.2857    & 0.2571    & \textbf{0.2286}    & \textbf{0.2286} \\ 
12&ALBERT       &ComplEx        &emb\_concat & \textbf{0.1821}  & \textbf{0.6196}  & \textbf{0.3361}    & \textbf{0.3098}    & \textbf{0.2857}    & 0.2571    & 0.0929 \\\hline
13&RoBERTa& -             & -         & 0.1011  & 0.3158  & 0.2286    & 0.2000    & 0.1429    & 0.1286    & 0.0929 \\
14&RoBERTa&Wikepedia2Vec  &emb\_concat & 0.1453  & 0.4176  & \textbf{0.3361}    & 0.2857    & 0.2571    & 0.2000    & \textbf{0.2286} \\ 
15&RoBERTa      &ComplEx        &emb\_concat & 0.1660  & 0.5174  & \textbf{0.3361}    & \textbf{0.3098}    & 0.2000    & 0.2571    & \textbf{0.2286} \\\hline
\hline
\multicolumn{11}{|c|}{Experimental results using CLEF'2020 CheckThat!\ dataset} \\\hline
16&SVM(TF.IDF)       & -            & -            & 0.0946    & 0.1531    & 0.0000    & 0.0600    & 0.0400    & 0.0450    & 0.0240   \\ 
17&SVM(TF.IDF)       & ComplEx      &emb\_concat    & 0.0923    & 0.1170    & 0.0000    & 0.0200    & 0.0500    & 0.0675    & 0.0270   \\ 
18&BiLSTM+att        & -        & -             & 0.0151    & 0.0320    & 0.0000    & 0.0100    & 0.0150    & 0.0075    & 0.0090   \\ 
19&BiLSTM+att        & ComplEx   &emb\_concat    & 0.0183    & 0.0320    & 0.0000    & 0.0200    & 0.0100    & 0.0100    & 0.0090   \\ 
20&BERT              & -        & -            & 0.0262    & 0.0819    & 0.0500    & 0.0300    & 0.0250    & 0.0125    & 0.0110 \\
21&BERT              & ComplEx   &emb\_concat    & 0.0373    & 0.0819    & 0.0500    & 0.0500    & 0.0350    & 0.0175    & 0.0130 \\ 
22&ALBERT            & -        & -            & 0.0537    & 0.2145    & 0.2000    & 0.0800    & 0.0500    & 0.0250    & 0.1600 \\
23&ALBERT            & ComplEx   &emb\_concat    & \textbf{0.1036}  & \textbf{0.2644}    & \textbf{0.2500}   & \textbf{0.0900}  & \textbf{0.0550}  & \textbf{0.0275}  & \textbf{0.0170} \\ 
24&RoBERTa           & -         & -            & 0.0424    & 0.1315    & 0.1000    & 0.6000     & 0.0400      & 0.0200    & 0.1400 \\
25&RoBERTa           & ComplEx   &emb\_concat    & 0.0923  & 0.1814    & 0.1500   & 0.0700  & 0.0450  & 0.0225  & 0.0150 \\ 
\hline
\end{tabular}
\end{table*}

\begin{table}
\footnotesize
\centering
\caption{Descriptive analysis of the test set of 2020 dataset. Note, this table consist of only check-worthy sentences (denoted as CW). The results we investigate here is obtained using ALBERT language model, ComplEx entity embedding method, and emb\_comcat method.}
\begin{tabular}{|l|r|r|r|r|r|r|} \hline
debate type   & \# of transcript & \# of cw & cw/transcript & \# of entities/CW & Recall (classification)  \\ \hline
Democratic    & 4   & 26    & 6.5   & 2.62  & 0.31  \\ 
Republican    & 1   & 7     & 7     & 2     & 0.15\\
Mixed~        & 2   & 23    & 11.5  & 2.57  & 0.17\\
Trump alone   & 13  & 83    & 6.38  & 1.16  & 0.11 \\ \hline
\end{tabular}
\label{tab:analysis}
\end{table}

\begin{table}
\footnotesize
\centering
\caption{A selected cases of check-worthy sentences, the identified entities, and if ALBERT + ComplEx successfully identified it as check-worthy. \textbf{Bold} denotes the identified entities.}
\begin{tabular}{|l|U|Z|Z|} \hline
Speaker & Sentence & \# of entities & Predicted correctly \\ \hline
Trump   & \textbf{Trump} was totally against the war in \textbf{Iraq}.  & 2 &  Y \\ \hline
Trump   & But when you make your car or when you make your \textbf{air conditioner}, and you think you're going to fire all of our workers and open up a new place in another country, and you're going to come through what will be a very strong border, which is already -- you see what's happened; 61 percent down now in terms of illegal people coming in. & 1 & N \\ \hline
Cruz    & \textbf{Bernie} helped write \textbf{Obamacare}.  & 2 & Y \\ \hline
Cruz    & There are many people in \textbf{America} struggling with exactly what you are, in the wreckage of \textbf{Obamacare}, with skyrocketing premiums, with deductibles that are unaffordable, and with really limited care. &  2 & N  \\ \hline
Clinton & \textbf{Trump}’s on record extensively supporting intervention in \textbf{Libya}, when \textbf{Gadhafi} was threatening to massacre his population. & 3 & Y \\ \hline
Clinton & And I do think there is an agenda out there, supported by my opponent, to do just that.  & 0 & N \\ \hline
\end{tabular}
\label{tab:cases}
\end{table}

\subsection{Failure Analysis}
\label{sec:5.5}
Due to the overall low performance of our models, we aim to identify the bottleneck of our framework, on the task of check-worth sentences identification. 

Table~\ref{tab:analysis} shows that there are differences numbers of transcript and check-worthy sentences from different parties that participated in the debate. That is, check worthy sentences from interviews and speeches given by Trump alone makes up as much as 60\% of the total number of check-worthy sentences. Moreover, there is a noticeable difference between transcript including Democrat candidates and Republic candidates. For example, check-worthy sentences from Democratic debates have a much higher number of entities detected per check-worth sentences, than those from Republican candidates. Further, we notice a strong correlation between the number of entities per check-worth sentences and the recall from the classification task, i.e., Democratic debates has an average number of entities per check-worthy sentences of 2.62, and has a recall of 0.31, compared to the republican debate that has 2 entities per check-worth sentences and only 0.15 recall, and transcripts considering Trump alone has an average number of entities as low as 1.16, with a recall of 0.11. 

Moreover, Table~\ref{tab:cases} shows 6 sentences, with number of identified entities, and if the classifier correctly identified the sentence as check-worthy. We observe that in the three of false negative cases (row 2,4, and 6), the number of entities are either less than or equals to 2. 

We therefore postulate that the number of entities per sentence can indeed affect the performance of our proposed framework.


\subsection{Recap of Main Findings}
\label{sec:5.6}

In this section, we recap on our main findings for RQs 1-4 and indicate the implications of our study, by use of Tables~\ref{tab:classification-summary} \& \ref{tab:ranking-summary}. For each language model, the summarising tables present results obtained using three conditions: language model only; with entity pair representation using Wikipedia2Vec KG embedding model and using emb\_concat method; and with entity pair representation using Compl\-Ex embedding and using emb\_concat method. We do not include emb\_prod method in our summarising tables, as our results for RQ3 showed that  emb\_concat consistently outperforms emb\_prod across both 2019 \& 2020 datasets and both classification and ranking tasks (see Section~\ref{sec:5.3}). 

For the classification task (Table~\ref{tab:classification-summary}, on the 2019 dataset), we highlight our conclusion from RQ1 (see Section~\ref{sec:5.1}) that the ALBERT language model (rows 11 - 13) significantly outperforms all other language models. 
We also confirm our conclusion from RQ2 (see Section~\ref{sec:5.2}) that entity embeddings improve language models' performance at identifying check-worthy sentences (row 3 \& 4 vs.\ 2, rows 6 \& 7 vs.\ 5, rows 9 \& 10 vs.\ 8, rows 12 \& 13 vs.\ 11, rows 15 \& 16 vs.\ 14).
Finally, we highlight our conclusion from RQ4 (see Section~\ref{sec:5.4}) that the ComplEx embedding method (rows 4, 7, 10, 13 \& 16) -- which uses the facts-alone KG embedding -- significantly outperforms the semantic KG embedding method (i.e., Wikipedia2Vec, rows 3, 6, 9, 12 \& 15). 

For the ranking task (Table~\ref{tab:ranking-summary}, on both the 2019 \& 2020 datasets), we draw similar conclusions as for classification task: The ALBERT language model (rows 10 - 12 for the 2019 dataset, rows 22 \& 23 for the 2020 dataset) consistently outperforms all other language models, while using the Compl\-Ex embedding model (rows 3, 6, 9, 12, 15 for the 2019 dataset, 17, 19, 21, 23, 25 for the 2020 dataset) consistently outperforms all other KG embedding models. Moreover, ALBERT + ComplEx + emb\_concat (row 11 for the 2019 dataset, row 20 for the 2020 dataset) obtains the best performance among all tested models. Thus, we conclude that ALBERT, ComplEx, and emb\_concat, can best identify and rank check-worthy sentences in a given speech or debate transcript.

In short, the findings of our study can thus be summarised as follows:
\begin{itemize}
    \item Deep neural language models help identify the sentences that require further manual fact-checking;
    \item Embedded entities within sentence help identify the sentences that requires further manual fact-checking;
    \item The most effective way to combine entity pair representation with text representation is to concatenate two vectors together;
    \item The tested facts-alone KG embedding models perform better than tested semantic KG embedding method (i.e., Wikipedia2Vec). The best performing KG embedding model in our study is the ComplEx model.
    \item The performance of our framework is affected by the number of entities present in the sentence. Since Donald Trump prefer sentences with less entities, it's difficult to identify his sentences as check-worthy than other candidates. 
\end{itemize}

\section{Conclusions} \label{sec:conclusion}
In this paper, we proposed a uniform framework for the task of check-worthy sentence identification, formulated as either a classification or a ranking task. We proposed to use BERT-related pre-trained language representations, and, in a novel manner, integrated entity embeddings obtained from knowledge graphs into the classifier and ranker. When considering the check-worthy sentence identification as a classification task, our experiments -- conducted using the CLEF'2019 and 2020 CheckThat!\ datasets -- showed that the ALBERT model outperforms the SVM(TF.IDF), BiLSTM+att, BERT, and RoBERTa text representation models. 
Moreover, the application of our proposed Entity-Assisted language models further improved the performance of the SVM(TF.IDF), BiLSTM+att, and all the BERT-related language models, over the models that combine language models with entity similarities and relatedness. When considered as a ranking task, we found that all types of entity embeddings improve all the language models in identifying the top ranked check-worthy sentences, but do not perform as well in the lower ranks.
Furthermore, on both classification and ranking tasks, using ALBERT for text representation consistently performs the best among all the tested text representation models, with or without entity features. In addition, we found that use of the DistMult and ComplEx KG embedding models both improve all the language models the most, while ALBERT + ComplEx achieved the best F1 on classification task on the 2019 dataset, and the best MAP performance on both the 2019 (a tie with the best performing submission) and 2020 datasets. 
Thus, we conclude that our framework, which combines deep learning language models with embedded entity representations in a novel manner, can achieve the state-of-the-art performance in identifying check-worthy sentences. Moreover, we argue that given the flexibility of our framework, achieved by concatenating sentence and entities representations instead of jointly training them, it can be applied to a number of sentence classification tasks, using an appropriate language model and an appropriate KG embedding model. Our work also gives support to future workflows where journalists and representatives of other fact-checking organisations could benefit from accurate assistive classifiers, to focus their efforts only on a subset of suspicious claims that are worth checking thereby ensuring a faster and wider dissemination of news. Our future research plan regarding the check-worthiness task are 2-fold: first, we plan to enrich the information of the identified entities, (e.g., by adding the type of the entity, and the place and time of the entity, if applicable); second, we aim to research new approaches leveraging social media in check-worthy sentences classification and retrieval. 


\bibliographystyle{unsrt}  
\bibliography{templateArxiv}

\end{document}